\documentclass[lettersize,journal,twoside]{IEEEtran}
\usepackage{amsmath,amsfonts}
\usepackage{algorithmic}
\usepackage{algorithm}
\usepackage{array}
\usepackage[caption=false,font=normalsize,labelfont=sf,textfont=sf]{subfig}
\usepackage{textcomp}
\usepackage{stfloats}
\usepackage{url}
\usepackage{verbatim}
\usepackage{graphicx}
\usepackage{cite}
\usepackage{booktabs}
\usepackage[dvipsnames,table]{xcolor}
\usepackage{colortbl}      
\usepackage{amssymb}
\usepackage{ulem}
\usepackage{booktabs}
\usepackage{amsfonts}
\usepackage{nicefrac}
\usepackage{microtype}
\usepackage{enumitem}
\usepackage{multirow}
\usepackage{amsmath}
\usepackage{mdframed}
\usepackage[
    colorlinks=true,
    linkcolor=blue,
    citecolor=blue,
    urlcolor=blue
]{hyperref}
\makeatletter
\def\@cite#1#2{{\color{blue}[{#1\if@tempswa , #2\fi}]}}
\makeatother
\begin{document}

\title{Open-UniMo: Towards Unified Motion-Language Understanding and Generation in the Open World}

\author{Guocun Wang,
        Kenkun Liu,
        Guorui Song,
        Jing Lin,
        Zhe Huang,
        Luyuan Zhang,
        Dake Zhong, 
        Choo Sin Wai,\\
        Xiaoguang Han,~\IEEEmembership{Member,~IEEE},
        and Haoqian Wang,~\IEEEmembership{Member,~IEEE}%
        
\thanks{This work was supported by the Shenzhen Science and Technology Project under Grant KJZD20240903103210014.
\textit{(Guocun Wang, Kenkun Liu, and Guorui Song are co-first authors.) (Corresponding author: Haoqian Wang.)}}
\thanks{Guocun Wang, Guorui Song, Zhe Huang, Luyuan Zhang, Dake Zhong, Choo Sin Wai, and Haoqian Wang are with Tsinghua University, Beijing 100190, China
(e-mail: \{wang-gc25, sgr24, huangz23, louyuan-24, zdk25, caosw25\}@mails.tsinghua.edu.cn; wanghaoqian@tsinghua.edu.cn).}
\thanks{Kenkun Liu and Xiaoguang Han are with SSE, The Chinese University of Hong Kong, Shenzhen 518172, China
(e-mail: kenkunliu@link.cuhk.edu.cn, hanxiaoguang@cuhk.edu.cn).}
\thanks{Jing Lin is with MMLab, Nanyang Technological University, Singapore 639798
(e-mail: jing026@e.ntu.edu.sg).}
}

\IEEEaftertitletext{%
\begin{center}
    \includegraphics[width=0.98\textwidth]{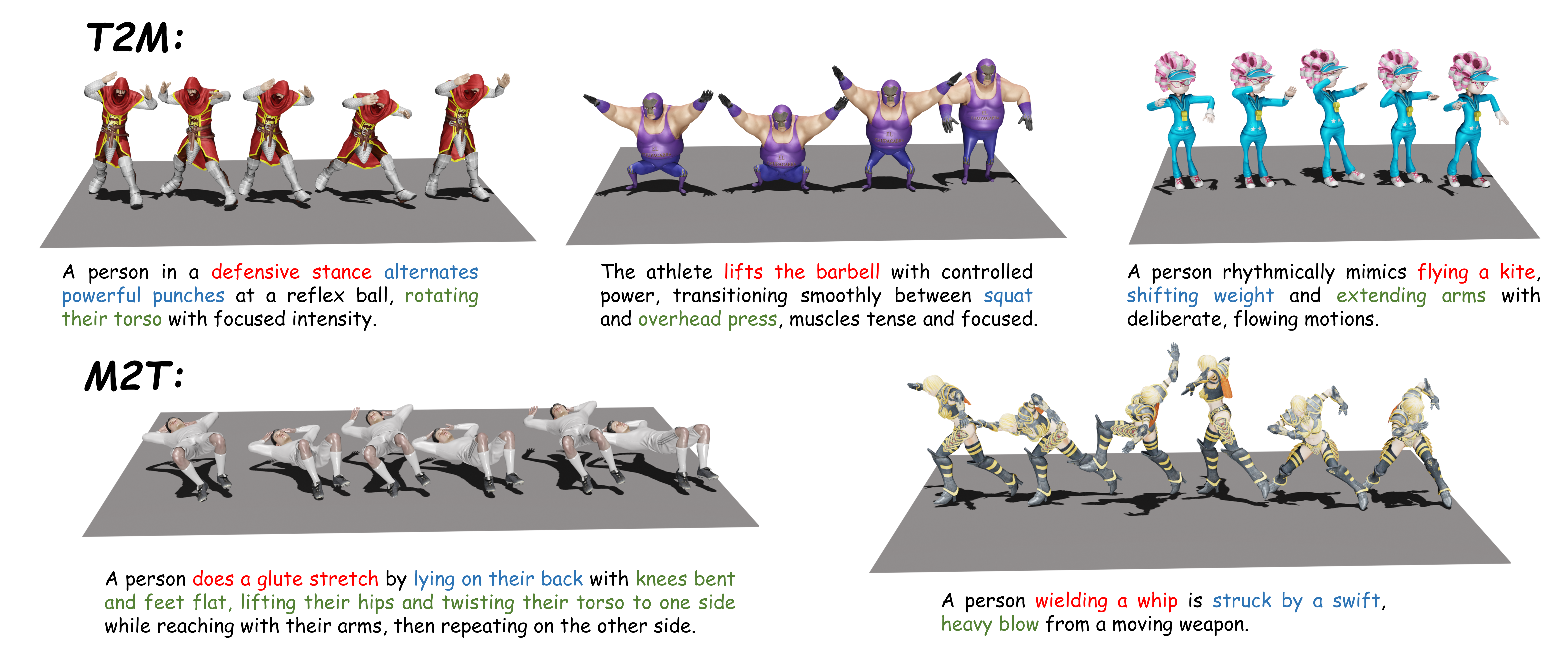}
    \refstepcounter{figure}
    \label{fig:teaser_image}
    
    \vspace{0.3em}
    {\footnotesize Fig.~\thefigure. Overview of Open-UniMo, a unified framework that supports text-to-motion generation and motion-to-text understanding in open-world scenarios.}
\end{center}
}

\maketitle

\begin{abstract}
Unified motion generation and understanding is crucial for embodied AI systems that can both synthesize and interpret human actions in open-world environments. Existing motion-language models often treat motion as an auxiliary modality of a language model, leading to text-dominated representations and limited cross-modal interaction. Moreover, the next-token prediction paradigm is not naturally suited to long motion sequences, where autoregressive generation may accumulate prediction errors. To address these challenges, we propose Open-UniMo, a unified Large Motion-Language Model (LMLM) trained on million-scale open-world motion-language data. Open-UniMo promotes modality parity by extending Qwen's vocabulary of about 150K text tokens with 64K motion tokens, enabling motion and language to share a unified token space. We further introduce motion-consistent Chain-of-Thought (CoT) reasoning as an intermediate representation to bridge language semantics and motion dynamics. Open-UniMo is trained with a two-stage pipeline, where supervised fine-tuning (SFT) establishes CoT-guided bidirectional motion-language mapping and Group Relative Policy Optimization (GRPO) improves semantic alignment while mitigating cumulative errors in autoregressive motion-token generation. To support comprehensive evaluation, we propose Open-MoBench, a VLM-guided benchmark for assessing text-to-motion (T2M) generation, motion-to-text (M2T) understanding, and bidirectional consistency. Extensive experiments show that Open-UniMo achieves state-of-the-art performance on both conventional metrics and Open-MoBench. Furthermore, ablation studies reveal that M2T understanding is not primarily limited by motion-token vocabulary size; instead, coupling M2T with the learnable T2M generation path yields stronger cross-modal representations, demonstrating that generation can facilitate understanding in AR-based motion-language modeling. The code will be released at \url{https://github.com/GuocunWang/Open-UniMo}.
\end{abstract}


\section{Introduction}

\IEEEPARstart{H}{uman} motion generation and motion understanding are long-standing problems in computer vision and graphics, enabling applications in animation, gaming, AR/VR, controllable video generation, and sports analysis~\cite{zhu2024champ,hu2024animate,qiu2025lhm,wang2024disco,song2025towards}. Motion generation aims to synthesize realistic movements that follow textual instructions, while motion understanding seeks to interpret motion sequences and produce coherent descriptions. Although these two tasks are inherently inverse, they have been largely studied in isolation~\cite{MLD,MotionDiffuse,Momask,T2M,T2M-GPT,MotionAgent,MotionR1}, overlooking their bidirectional synergy. Recent efforts attempt to unify text-to-motion (T2M) generation and motion-to-text (M2T) understanding within a single large language model (LLM) backbone~\cite{Mg-motionllm,M3gpt,MotionGPT,Motiongpt2,Motiongpt3}. However, existing unified models still lag behind task-specific methods, especially when moving from controlled benchmarks to open-world motion-language data.

This gap stems from several fundamental challenges. First, motion-language alignment is intrinsically weak. Textual instructions are often simple and coarse, while motion contains rich spatiotemporal details, leading to an intrinsically one-to-many correspondence that makes it difficult to learn fine-grained control from paired supervision alone. Second, the next-token prediction (NTP) paradigm is ill-suited for motion. Unlike natural language with strong sequential causality, human motion is a continuous temporal process with weaker causality; autoregressive token prediction therefore suffers from cumulative errors, especially for long sequences. Third, LLM-based unification suffers from modality imbalance. Many motion-language models treat motion as a secondary modality injected into an LLM, producing a text-dominated embedding space and restricting cross-modal interaction~\cite{MotionAgent,MotionGPT}. This issue becomes more pronounced in open-world settings: large-scale datasets such as MotionMillion~\cite{MotionMillion} cover substantially more diverse motion dynamics and linguistic patterns, and insufficient motion-token capacity may increase reconstruction errors that further propagate to downstream T2M generation, consistent with observations in large-scale motion generation work such as ScaMo~\cite{Scamo}. Finally, evaluation remains limited. Traditional metrics (e.g., FID, R-Precision, BLEU) focus on surface-level similarity or lexical overlap and fail to assess semantic alignment, temporal coherence, physical plausibility, and robustness under one-to-many correspondence, while human ratings are costly and difficult to scale.

A key question for building an LMLM is how much discrete capacity should be allocated to the motion modality. Existing AR-based motion-language models~\cite{T2M-GPT,MotionGPT,Motiongpt2,MotionAgent,MotionR1} usually rely on hundreds of motion tokens, which may be adequate for HumanML3D~\cite{T2M} but insufficient for open-world motion distributions. Even in AR-based visual generation, representative methods~\cite{LlamaGen,VAR,Chameleon,Liquid,wang2024emu3,Emu3_5,chen2025janus,Omnigen-ar} often use thousands to tens of thousands of visual tokens, far more than the motion tokens used in prior AR-based motion-language models. Inspired by recent large-scale motion tokenization studies such as GoToZero~\cite{MotionMillion} and ScaMo~\cite{Scamo}, which show the importance of scaling motion tokens for open-world T2M generation, we explore a substantially larger motion vocabulary by introducing 64K motion tokens into an LLM vocabulary. This design promotes modality parity by treating motion as a first-class token modality, and further enables us to examine how high-capacity motion tokenization interacts with unified motion generation and understanding under an autoregressive framework.

To address these challenges, we present \textbf{Open-UniMo}, a unified Large Motion-Language Model (LMLM) for open-world motion generation and understanding. Open-UniMo is built upon Qwen2.5-3B-Instruct~\cite{qwen2_5} and extends its vocabulary of about 150K text tokens with 64K motion tokens, forming a unified token space that treats motion as a first-class modality rather than an auxiliary side channel. To bridge language semantics and motion dynamics, we introduce motion-consistent Chain-of-Thought (CoT) reasoning as an intermediate representation. Unlike caption-rewrite approaches such as MotionR1~\cite{MotionR1}, which construct CoT traces only from text and may introduce hallucinated reasoning, we render motions into videos with Blender and use Qwen2.5-VL-72B~\cite{qwen2_5_vl} to generate CoT traces grounded in actual motion dynamics. Open-UniMo is then trained with a two-stage pipeline. In the supervised fine-tuning (SFT) stage, the model learns CoT-guided bidirectional motion-language mapping. In the second stage, Group Relative Policy Optimization (GRPO)~\cite{GRPO} further refines the model with task-specific rewards for format correctness and cross-modal alignment, improving robustness beyond token-level supervised learning.

To comprehensively and reproducibly evaluate unified motion-language models, we introduce \textbf{Open-MoBench}, a VLM-guided evaluation framework that scores both T2M and M2T outputs across multiple diagnostic dimensions. Open-MoBench evaluates T2M generation along physical plausibility, semantic alignment, and temporal coherence, and evaluates M2T understanding along detail consistency, semantic alignment, and temporal coherence. Beyond single-direction evaluation, it also introduces a bidirectional consistency protocol that exploits the intrinsic duality between T2M and M2T: a unified model should reconstruct semantically equivalent descriptions under $T \rightarrow \hat{M} \rightarrow \hat{T}'$ and preserve motion semantics under $M \rightarrow \hat{T} \rightarrow \hat{M}'$. This cyclic evaluation serves as a principled test of whether a unified model preserves cross-modal semantics across bidirectional transformations, rather than merely learning one-way correlations between motion and language. We further validate the reliability of Open-MoBench through a VBench-style~\cite{vbench} human preference study in Sec.~\ref{sec:human_preference_testing}.

Compared with our conference version UniMo~\cite{unimo}, the major contributions and extensions are summarized as follows:

\begin{itemize}

\item We introduce Open-UniMo, a unified LMLM for open-world T2M and M2T. Compared with UniMo, Open-UniMo scales unified modeling to million-scale motion-language data and extends Qwen's vocabulary with 64K motion tokens, treating motion as a first-class modality.

\item We scale and validate the CoT-guided SFT$\rightarrow$GRPO training pipeline under open-world motion-language distributions. With CoT as a shared intermediate representation and GRPO-based post-training, this pipeline improves semantic alignment and mitigates cumulative errors in autoregressive motion-token prediction.

\item We propose Open-MoBench, a VLM-guided benchmark that jointly evaluates T2M generation, M2T understanding, and bidirectional consistency across diverse dimensions, enabling scalable and reproducible assessment of unified motion-language intelligence.

\item We experimentally show that T2M-only training can be effectively optimized, whereas M2T-only training does not show a consistent positive trend with token capacity, indicating that M2T understanding is not primarily limited by motion-token vocabulary size. Unified training further couples M2T with the learnable T2M path, demonstrating that generation facilitates understanding in AR-based motion-language models.

\end{itemize}

\section{Related Work}
\subsection{Human Motion Generation}

Text-conditioned human motion generation has progressed through several methodological stages.
Early attempts~\cite{language2pose,ghosh2021synthesis,huang2020dance,lin2018generating,plappert2018learning} typically relied on deterministic mappings from text or action labels to pose sequences.
While conceptually straightforward, these methods often suffered from limited diversity and weak generalization.
With the development of probabilistic modeling and text-motion datasets, methods such as T2M~\cite{T2M} advanced language-conditioned motion synthesis, while TM2T~\cite{TM2T} further explored reciprocal text-to-motion and motion-to-text generation.
In parallel, generative adversarial frameworks~\cite{cai2018deep,wang2020learning} explored action-conditioned motion synthesis.

More recently, diffusion-based generative modeling~\cite{ho2020denoising,nichol2021improved,song2020denoising} has reshaped the field.
Representative works such as MDM~\cite{MDM}, MotionDiffuse~\cite{MotionDiffuse}, and MLD~\cite{MLD} adapt diffusion models to text-to-motion generation, while subsequent studies improve fine-grained semantic control~\cite{Fg-T2M++,Finemogen} and multi-motion generation~\cite{M2d2m}.
Beyond diffusion, MoMask~\cite{Momask} introduces masked motion token prediction, and another line of work adopts autoregressive generation over discrete motion tokens~\cite{T2M-GPT,MotionGPT,Motiongpt2,MotionChain,MotionAgent,MotionR1}.
Recent studies~\cite{Scamo,MotionMillion,vimogen,cao2025being,HY-Motion} further push motion generation toward open-world and generalizable settings. ScaMo~\cite{Scamo} studies scaling laws in AR-based motion generation, GoToZero~\cite{MotionMillion} introduces million-scale motion-language data for zero-shot generalization, and ViMoGen~\cite{vimogen} explores generalizable motion generation from data, model, and evaluation perspectives.
Despite these advances, most methods focus on T2M generation alone, and AR-based approaches still face challenges in long-horizon temporal coherence and fine-grained semantic alignment, partly due to cumulative errors in next-token prediction over motion-token sequences.

\subsection{Unified Motion Generation and Understanding}
Unified multimodal large language models (LLMs) have recently achieved remarkable progress in bridging understanding and generation in vision~\cite{xie2024show,chen2025janus,wang2024emu3,pan2025transfer,deng2025emerging,chen2025blip3,Emu3_5,Omnigen-ar,OmniGen2}.
Systems such as GPT-4o, EMU-3~\cite{wang2024emu3}, and Bagel~\cite{deng2025emerging} demonstrate the trend of integrating understanding, reasoning, and generation within unified multimodal interfaces, improving controllability and interaction in many scenarios.

Motivated by these advances, several studies attempt to unify human motion generation and understanding within a single LLM-based framework~\cite{Motiongpt3,MotionAgent,MotionGPT,zhang2024motiongpt,unimo}.
These approaches typically encode motion into discrete tokens or continuous features and integrate them with an LLM for bidirectional motion-language mapping.
MG-MotionLLM~\cite{Mg-motionllm} introduces multi-granularity training tasks to improve motion comprehension and generation at different temporal and semantic levels.
MotionGPT-3~\cite{Motiongpt3} treats motion as a second modality and uses modality-specific modeling to reduce cross-modal interference.
UniMo~\cite{unimo} introduces CoT and GRPO-based post-training for motion generation and understanding.
Beyond text-motion pairs, M$^3$GPT~\cite{M3gpt} unifies text, music, and motion within a multimodal and multitask framework.

Despite these advances, open-world motion-language unification remains challenging.
Most existing models are still evaluated on controlled benchmarks~\cite{T2M,KIT-ML} and rely on limited-capacity motion representations.
As a result, the role of high-capacity motion tokens and the interaction between AR-based generation and understanding remain insufficiently explored.
In contrast, Open-UniMo expands the motion vocabulary to 64K tokens, trains on million-scale open-world data, and analyzes how the learnable T2M generation path benefits M2T understanding in unified AR-based modeling.

\subsection{Chain-of-Thought for Generative Models}
Chain-of-Thought (CoT) reasoning has been shown to enhance the reasoning capability and interpretability of LLMs by decomposing complex problems into structured intermediate steps~\cite{wei2022chain,Deepseek-r1}.
Recent generative models further extend this idea beyond textual reasoning by introducing explicit planning or reasoning traces before synthesis~\cite{deng2025emerging,guo2025can,Got-r1,Imagegen-cot,T2i-r1}, which improves controllability for complex prompts and provides more interpretable generation processes.

In human motion generation, MotionR1~\cite{MotionR1} introduces decomposed CoT reasoning and reinforcement learning (RL) for T2M.
Our conference version UniMo~\cite{unimo} constructs motion-consistent CoT by rendering motions into videos and using a VLM to generate reasoning traces grounded in actual motion dynamics, rather than relying only on caption rewriting.
Building on this formulation, Open-UniMo scales motion-consistent CoT to million-scale open-world data and uses it as a shared intermediate representation for both T2M and M2T.

\subsection{Evaluation of Motion-Language Models}
Existing motion-language evaluations mostly rely on conventional task-specific metrics, including FID and R-Precision for T2M generation~\cite{T2M}, as well as BLEU~\cite{Bleu} and BertScore~\cite{Bertscore} for M2T captioning. While useful for standardized comparison, these metrics provide limited diagnosis of semantic alignment, physical plausibility, temporal coherence, and bidirectional consistency.
Recent works have introduced more diagnostic protocols for motion generation. MotionMillion-Eval~\cite{MotionMillion} evaluates zero-shot T2M generation with human verification along text alignment, motion smoothness, and physical plausibility, but it is limited to the T2M direction and relies on manual assessment, making it difficult to scale. ViMoGen~\cite{vimogen} further proposes MBench, a hierarchical benchmark for generalizable motion synthesis across motion generalization, motion-condition consistency, and motion quality. However, these benchmarks still focus on generation quality and do not evaluate M2T understanding or bidirectional motion-language consistency.

Beyond motion, recent image and video generation benchmarks also suggest a shift from aggregate scores to fine-grained diagnostic evaluation. GenEval~\cite{geneval}, GenAI-Bench~\cite{genai-bench}, and WISE~\cite{wise} evaluate compositional alignment, complex prompt following, and world-knowledge-informed semantics, respectively, while VBench~\cite{vbench} decomposes video generation quality into multiple spatial and temporal dimensions. 
Following this diagnostic evaluation trend, Open-MoBench targets the missing setting in motion-language modeling by jointly evaluating T2M generation, M2T understanding, and bidirectional consistency with a scalable VLM-guided protocol, whose reliability is further validated through human preference testing.

\begin{figure*}[ht]
    \centering
    \includegraphics[width=\textwidth]{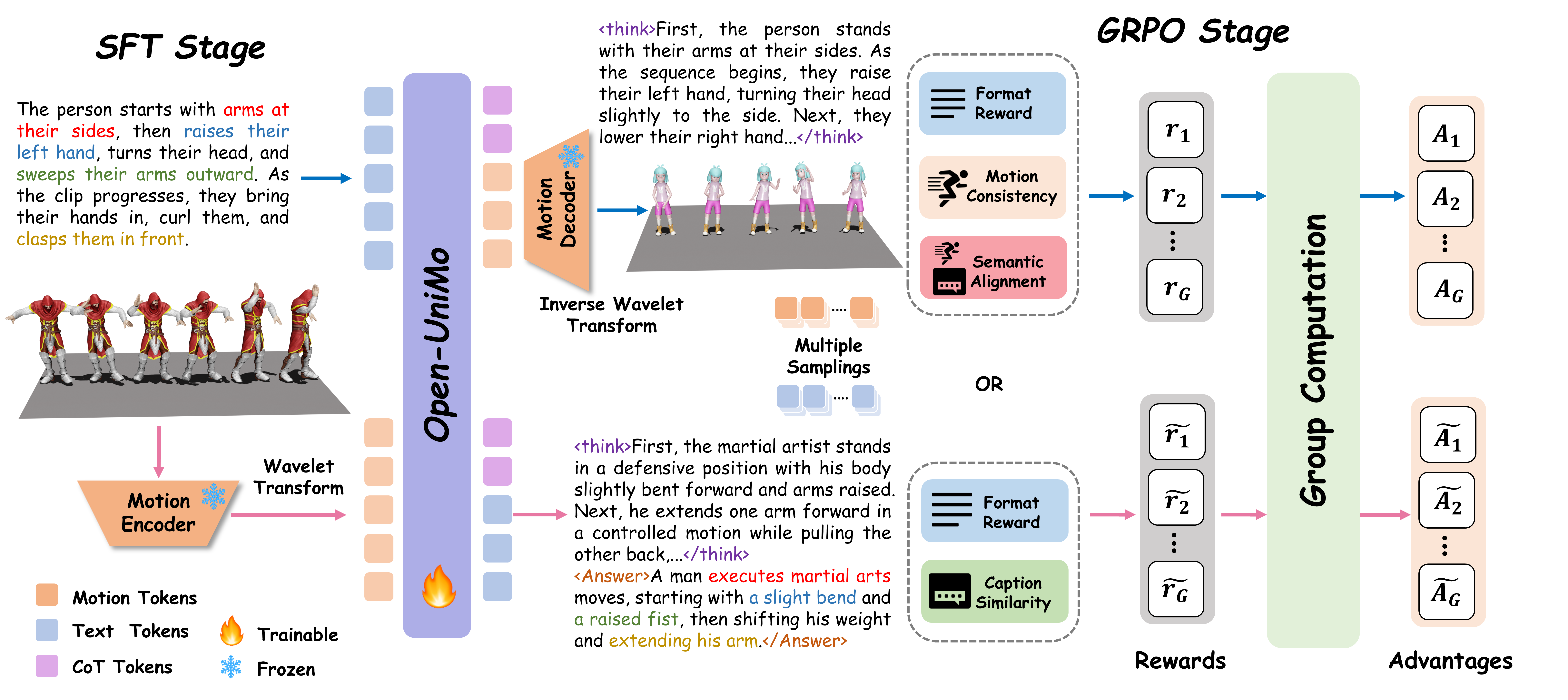}
    \caption{Overview of the Open-UniMo framework. Open-UniMo is trained in two stages with CoT reasoning. In the SFT stage, the model learns CoT-guided bidirectional motion-language mappings for both T2M and M2T tasks. In the GRPO stage, reinforcement learning further refines format compliance and motion-language alignment through multiple reward signals.}
    \label{fig:Open-UniMo}
\end{figure*}

\section{Method}
\subsection{Overview}
Our goal is to achieve open-world motion-language intelligence with a unified framework that supports both motion generation and motion understanding within a single Large Motion-Language Model (LMLM).
As shown in Fig.~\ref{fig:Open-UniMo}, we propose \textbf{Open-UniMo}, which treats motion and language with modality parity and enables bidirectional mappings between text and motion.
Open-UniMo is trained in two stages: (i) SFT with CoT, and (ii) RL using GRPO~\cite{GRPO} with task-specific rewards to improve semantic alignment and mitigate cumulative errors in next-token prediction for motion sequences.

\subsection{Motion Representation}
Following the motion representation used in GoToZero~\cite{MotionMillion}, we represent each frame as a refined 272-dimensional vector that avoids inverse-kinematics processing and supports lossless conversion to SMPL~\cite{loper2023smpl}.
Each pose is expressed as
\begin{equation}
x=\bigl\{\dot{r}^{x},\dot{r}^{z},\dot{r}^{a},j^{p},j^{v},j^{r}\bigr\}\in\mathbb{R}^{272},
\end{equation}
where $\dot{r}^{x},\dot{r}^{z}\in\mathbb{R}$ denote root linear velocities on the XZ-plane, $\dot{r}^{a}\in\mathbb{R}^{6}$ denotes root angular velocity in 6D representation, $j^{p},j^{v}\in\mathbb{R}^{3K}$ denote local joint positions and velocities, and $j^{r}\in\mathbb{R}^{6K}$ denotes local joint rotations in root coordinates, with $K$ being the number of body joints.

\subsection{Motion Tokenization}
To integrate motion into an autoregressive LLM, we discretize continuous motion sequences into motion token indices.
Following recent large-scale works~\cite{Scamo,MotionMillion}, we apply Finite Scalar Quantization (FSQ), a codebook-free quantization approach that discretizes latent vectors via deterministic rounding and avoids the storage overhead of VQ-VAE-style codebooks~\cite{T2M-GPT,MotionGPT,MotionAgent}.
To reduce quantization artifacts, we apply wavelet transforms before encoding and inverse transforms after decoding~\cite{MotionMillion}.

Given an encoder latent vector $z$, we normalize it into $[0,1]$, and quantize each dimension by rounding:
\begin{equation}
z_q = Q(z) = \mathrm{round}\bigl(f(z)\cdot(L-1)\bigr),
\end{equation}
where $L$ denotes the number of quantization levels per latent dimension, $d$ is the latent dimension, and $z_q\in\{0,1,\ldots,L-1\}^{d}$ is the resulting discrete FSQ code.
The tokenizer is trained with a reconstruction loss:
\begin{equation}
\mathcal{L}_{\mathrm{rec}} = \bigl\| m - \mathrm{Dec}(z_q) \bigr\|_2^2,
\end{equation}
where $m$ is the target motion and $\text{Dec}(z_q)$ denotes decoded motion from the quantized latent.
After training, the tokenizer is frozen during subsequent LMLM training.

\begin{figure}[htbp]
    \centering
    \includegraphics[width=\linewidth]{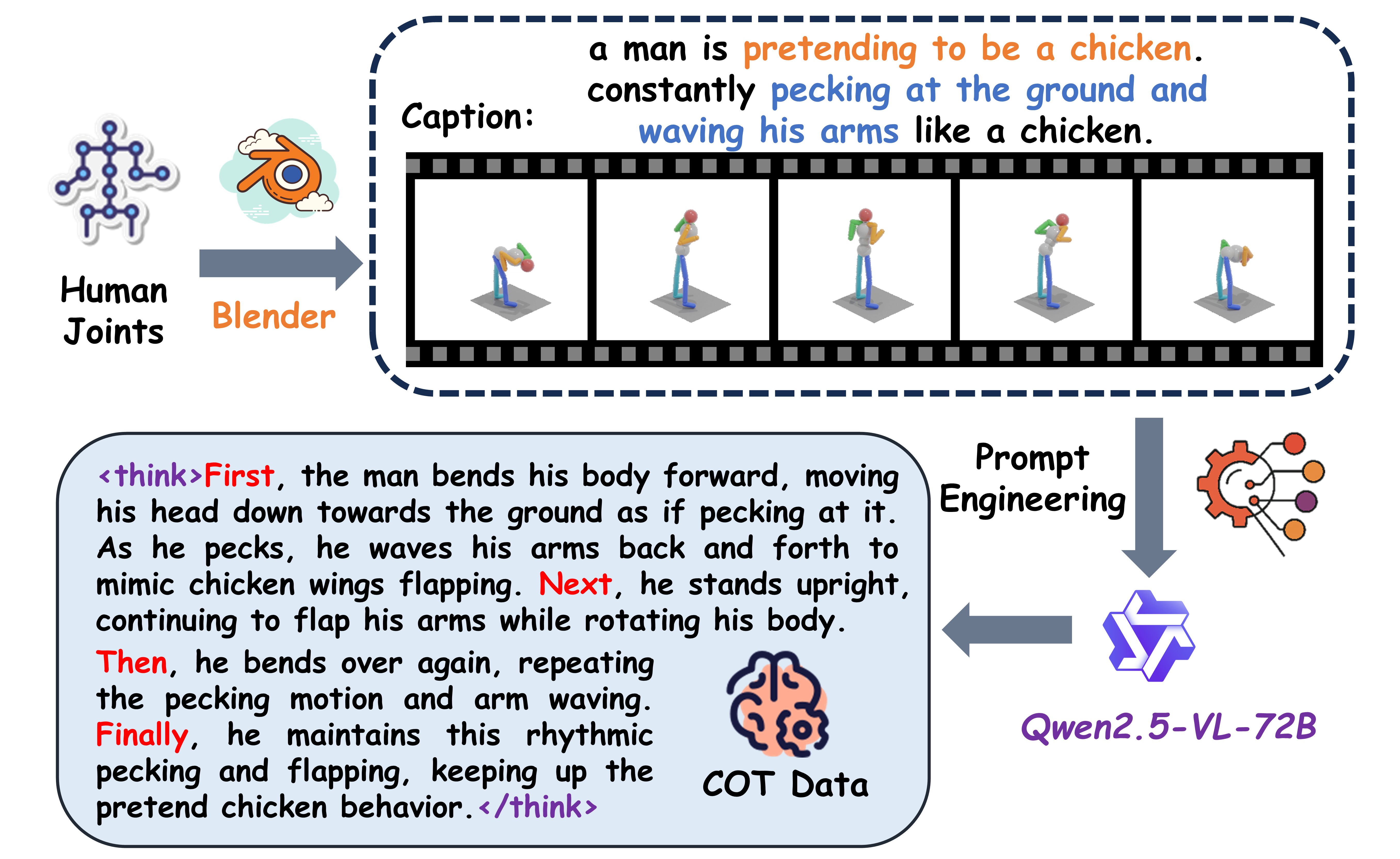}
    \caption{Illustration of the CoT annotation process. Human joint sequences are rendered in Blender and paired with captions, which are further processed by Qwen2.5-VL-72B to generate reasoning traces.}
    \label{fig:COT-Annotation}
\end{figure}

\subsection{Unified LMLM with Modality Parity}
Open-UniMo is built upon Qwen2.5-3B-Instruct~\cite{qwen2_5}.
To achieve modality parity, we extend the Qwen tokenizer vocabulary (about 150K text tokens) with an additional 64K motion tokens representing quantized motion primitives.
This design treats motion as a first-class modality with parity to language rather than an auxiliary side channel, enabling richer cross-modal interaction and providing sufficient discrete capacity for open-world motion distributions. Such capacity is important for large-scale motion modeling, since insufficient motion-token vocabularies can increase reconstruction errors and propagate them to autoregressive motion generation, as observed in ScaMo~\cite{Scamo}.
With a unified token space, Open-UniMo can perform both T2M and M2T by conditioning on the input modality and generating the target modality.

\begin{figure*}[ht]
    \centering
    \includegraphics[width=\textwidth]{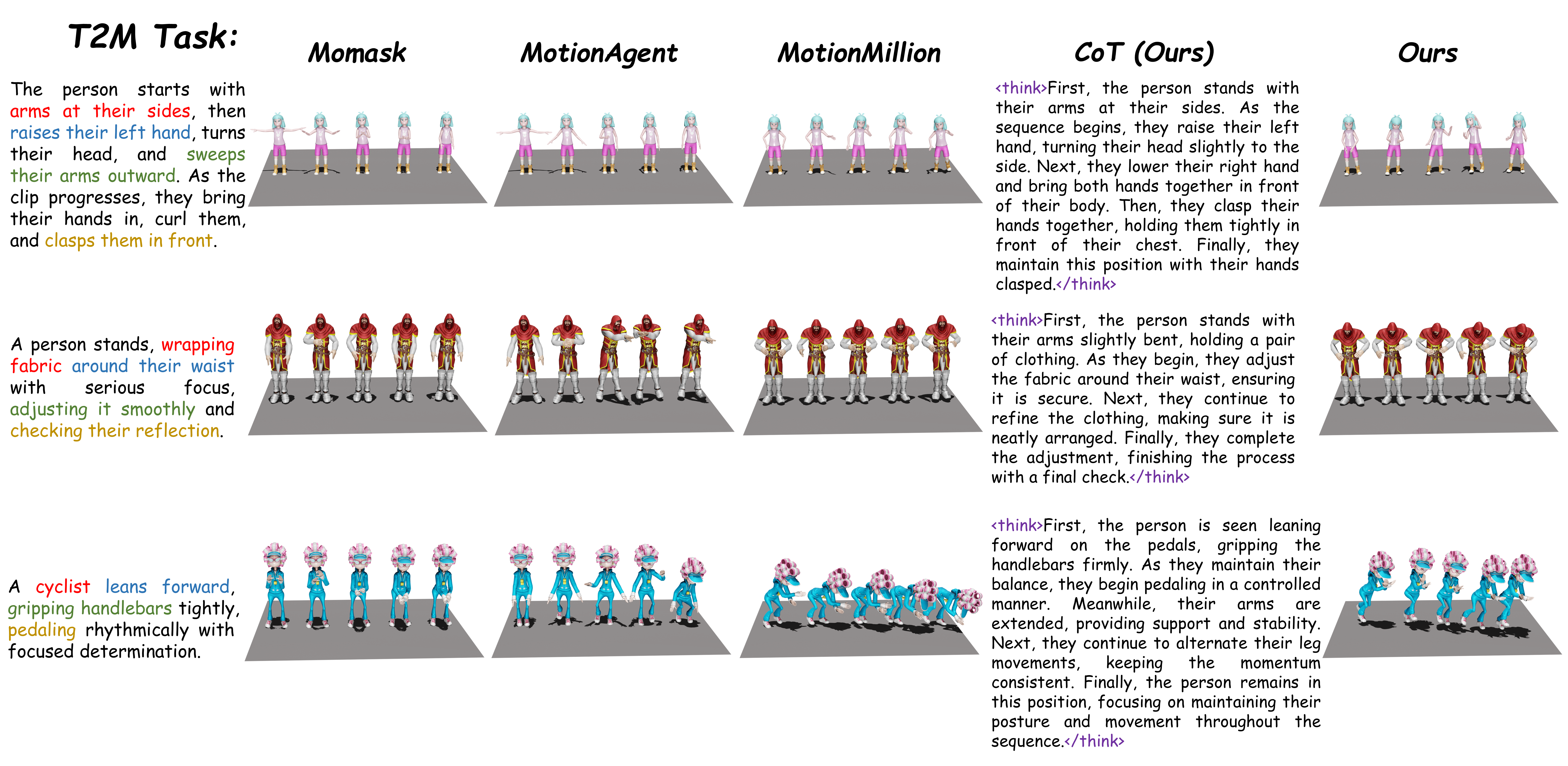}
    \caption{Qualitative comparison on Open-MoBench for the T2M task. Compared with existing representative approaches~\cite{MotionAgent,Momask,MotionMillion}, Open-UniMo generates motions that better capture semantic details and natural dynamics.}
    \label{fig:T2M}
\end{figure*}

\begin{figure*}[ht]
    \centering
    \includegraphics[width=\textwidth]{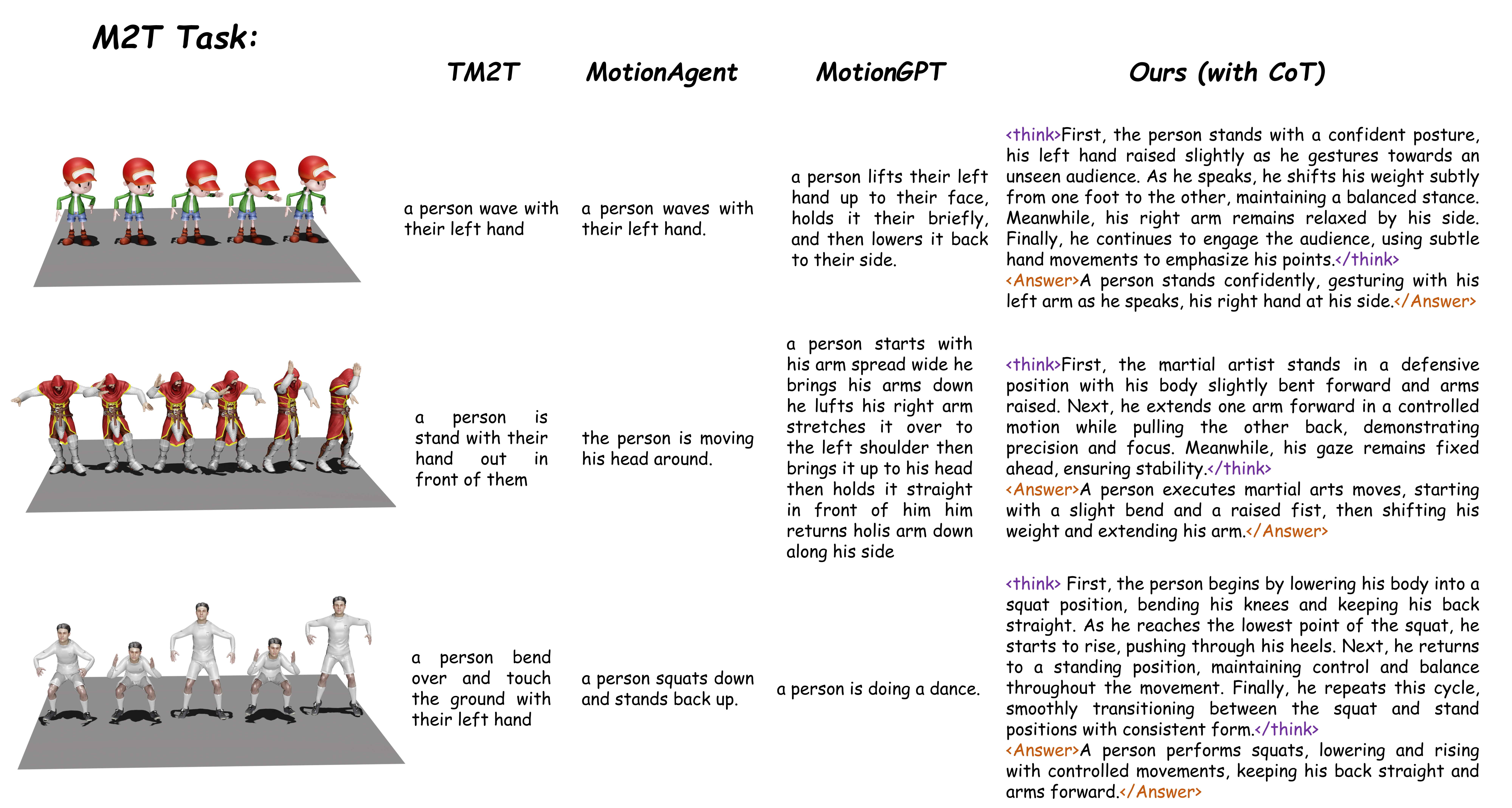}
    \caption{Qualitative comparison on Open-MoBench for the M2T task. Compared with open-source models~\cite{MotionGPT,MotionAgent,TM2T}, Open-UniMo generates descriptions that are more coherent, contextually relevant, and semantically faithful to the input motions.}
    \label{fig:M2T}
\end{figure*}

\subsection{Supervised Fine-Tuning with Chain-of-Thought}
A key challenge for unified motion-language modeling is bridging coarse text with rich motion details under one-to-many correspondence.
We adopt Chain-of-Thought (CoT) as an explicit intermediate representation that improves interpretability, enhances semantic alignment, and enables stepwise motion planning.
To obtain motion-consistent CoT, we follow UniMo's multimodal annotation strategy~\cite{unimo}.
As shown in Fig. \ref{fig:COT-Annotation}, we render motion sequences into videos via Blender and prompt Qwen2.5-VL-72B~\cite{qwen2_5_vl} with both the rendered video and its paired caption to generate step-by-step reasoning traces in the form \texttt{<think>...</think>}.
Compared to caption-rewrite CoT (e.g., MotionR1~\cite{MotionR1}), this grounded annotation reduces hallucination and ensures the CoT traces reflect actual motion dynamics.
In the SFT stage, the training samples encourage the model to think before answering. For the T2M task, the required format is \texttt{<think>\{CoT\}</think><Motion>\{Motion Tokens\}</Motion>}. For the M2T task, it is \texttt{<think>} \texttt{\{CoT\}</think><Answer>\{Caption\}</Answer>}.
Training on paired (motion, caption, CoT) triples provides semantic grounding and interpretability, and offers a strong initialization for bidirectional mapping before RL training.

\subsection{Reinforcement Learning with GRPO}
While CoT-guided SFT improves structure and alignment, autoregressive next-token prediction remains prone to cumulative errors in long motion sequences.
We therefore employ reinforcement learning with GRPO~\cite{GRPO}, a PPO-style algorithm that removes the need for an explicit critic by computing advantages through within-group normalization.

\paragraph{Group Relative Policy Optimization}
The GRPO~\cite{GRPO} algorithm enhances conventional policy-optimization by generating multiple candidate outputs per input and evaluating them collectively. Instead of depending on the critic network, it calculates advantage scores through normalization within the group to reduce variance in policy updates.
The objective is

\begin{eqnarray}
\mathcal{J}_{\text{GRPO}}(\theta) &=& E_c \left[ \frac{1}{G} \sum_{i=1}^{G} \min \left( 
\frac{\pi_{\theta}(o_i|q)}{\pi_{\text{old}}(o_i|q)} \hat{A}_i, \right. \right. \nonumber \\
&& \left. \left. \text{clip} \left( 
\frac{\pi_{\theta}(o_i|q)}{\pi_{\text{old}}(o_i|q)}, 1-\varepsilon, 1+\varepsilon \right) \hat{A}_i \right) \right] \nonumber \\
&& - \beta \cdot D_{\text{KL}}(\pi_{\theta} \parallel \pi_{\text{ref}}),
\end{eqnarray}
where \(\theta\) denotes the trainable policy parameters.  
For each query \(q\) the algorithm samples a fixed group of \(G\) candidate outputs \{\(o_{1},\dots,o_{G}\)\}. The term \(\pi_{\theta}(o_{i}|q)\) is the probability of generating output \(o_{i}\) under the current policy, while \(\pi_{\text{old}}(o_{i}|q)\) is the corresponding probability under the old policy.  
The clipping radius is \(\varepsilon\), and the coefficient \(\beta\) controls the strength of the KL penalty \(D_{\text{KL}}(\pi_{\theta}\parallel\pi_{\text{ref}})\), which regularizes the current policy toward a reference policy \(\pi_{\text{ref}}\).
The normalized group-relative advantage \(\hat{A}_{i}\) is defined as:
\begin{equation}
\hat{A}_i = 
\frac{r_i - \text{mean}(\{r_1,\dots,r_G\})}
{\text{std}(\{r_1,\dots,r_G\})},
\end{equation}
where $r_i$ is the reward assigned to output $o_i$. By normalizing rewards within each group, GRPO computes relative advantages, thereby reducing variance in the policy gradient estimate and promoting more stable optimization dynamics.

\paragraph{Unified Reward Formulation}
To balance structural correctness and semantic alignment during RL training, we design a composite reward scheme with task-specific components. The reward design is intentionally asymmetric because T2M and M2T optimize different information paths: T2M requires motion-space fidelity and motion-language alignment, whereas M2T requires semantic compression into language. Therefore, we define motion-space and cross-modal rewards for T2M, and a caption-level semantic reward for M2T. Each sample receives both a format reward for structural validity and task-specific alignment rewards, enabling GRPO to promote faithful motion generation and linguistically coherent captioning.

\textbf{Format Reward.}
To guide the model toward generating outputs that follow the predefined structural templates, we define a format reward~$R_{\text{format}}$.
For the T2M task, the template is
\verb|<think>{CoT}</think><Motion>{Motion Tokens}|
\verb|</Motion>|.
For the M2T task, it is
\verb|<think>{CoT}|
\verb|</think><Answer>{Caption}</Answer>|.
An exact match earns a reward of $0.5$; any deviation yields $0$.

\textbf{Modality Alignment Rewards.}
To measure the semantic correspondence between generated content and its reference, we employ a generalized cosine similarity function:
\begin{equation}
\mathcal{S}(x,y)=
\frac{x^\top y}{\|x\|_2\|y\|_2},
\end{equation}
where $x$ and $y$ are embeddings from frozen encoders~\cite{MotionMillion}.

For the T2M task, let $T$ denote the input caption, $M$ denote the reference motion, and $\hat{M}$ denote the generated motion. The motion consistency reward
$R_{\text{motion}}=\mathcal{S}(\phi_{\mathcal{M}}(\hat{M}),\phi_{\mathcal{M}}(M))$
encourages motion-space fidelity to the reference motion, while the semantic alignment reward
$R_{\mathrm{semantic}}=\mathcal{S}(\phi_{\mathcal{M}}(\hat{M}),\phi_{\mathcal{T}}(T))$
measures the consistency between the generated motion and the input caption. Here, $\phi_{\mathcal{M}}$ and $\phi_{\mathcal{T}}$ denote the frozen motion and text encoders from~\cite{MotionMillion}, respectively. The T2M reward design is inspired by~\cite{MotionR1}.
For the M2T task, let $\hat{T}$ denote the generated caption and $T$ denote the ground-truth caption. Instead of mirroring the T2M motion-space rewards, we use a caption similarity reward
$R_{\text{caption}}=\mathcal{S}(\psi_{\text{CLIP}}(\hat{T}),\psi_{\text{CLIP}}(T))$,
computed with CLIP's text encoder. This asymmetric design directly optimizes caption-level semantic consistency and is empirically supported by the weaker symmetric variant marked with $\triangledown$ in Tabs.~\ref{tab:t2m_results} and~\ref{tab:m2t_results}.

\textbf{Overall Reward Composition.}
The following formulas show the composition of both $R_{\text{T2M}}$ and $R_{\text{M2T}}$:
\begin{equation}
R_{\text{T2M}} = R_{\text{format}} + R_{\text{motion}} + R_{\text{semantic}},
\end{equation}
\begin{equation}
R_{\text{M2T}} = R_{\text{format}} + 2 \times R_{\text{caption}}.
\end{equation}

We scale $R_{\text{caption}}$ by $2$ to harmonize the magnitude of $R_{\text{M2T}}$ with $R_{\text{T2M}}$, ensuring balanced optimization across both tasks. 
By optimizing these rewards with GRPO, Open-UniMo improves semantic alignment and generation quality while mitigating cumulative errors in motion-token prediction.

\section{Open-MoBench}

\subsection{Overview}

Existing evaluations for motion-language models remain largely limited to conventional metrics, such as FID and R-Precision for T2M generation~\cite{T2M}, as well as BLEU~\cite{Bleu}, CIDEr~\cite{Cider}, and BertScore~\cite{Bertscore} for M2T captioning. While these metrics support standardized comparison, they provide limited diagnosis of semantic alignment, physical plausibility, temporal coherence, and cross-modal consistency. This limitation stems from the intrinsic one-to-many nature of motion–language correspondence: a single sentence, such as ``a person waves a hand happily,'' can correspond to multiple plausible motions differing in amplitude, rhythm, or style. Consequently, unidirectional metrics cannot fully reveal whether a model preserves semantics across modalities.

To address this gap, we propose \textbf{Open-MoBench}, a unified benchmark that adopts Gemini-3-Flash as the VLM evaluator to assess T2M generation, M2T understanding, and bidirectional consistency across multiple diagnostic dimensions. By jointly assessing these aspects, it provides a comprehensive foundation for open-world motion-language intelligence. We further validate the reliability of the VLM-as-Judge protocol through a VBench-style human preference study in Sec.~\ref{sec:human_preference_testing}, where model-level rankings induced by VLM scores are compared with rankings from human annotators.

\subsection{Text-to-Motion Evaluation}
Traditional T2M metrics focus on distributional similarity or retrieval accuracy, offering limited insight into whether generated motions truly reflect the intended semantics. To address this, Open-MoBench samples 1,000 text captions from two large open-world video datasets~\cite{llava-video-178k,Miradata}, covering diverse activities such as sports and daily actions. We further summarize the original video captions with Qwen2.5-72B-Instruct~\cite{qwen2_5} into motion-centric descriptions of approximately 20 words while preserving key action details. The vocabulary is rich and unconstrained, reflecting the complexity and variability of real-world motion descriptions. 
The generated motion is converted into SMPL~\cite{loper2023smpl} format and rendered into video using Blender. Each result is evaluated by VLM-as-Judge on a 0-5 scale across three dimensions:
\begin{itemize}
    \item \textbf{Physical Plausibility}
    examines biomechanical correctness and kinematic realism, including body coordination and ground contact consistency.
    
    \item \textbf{Semantic Alignment}
    assesses whether generated actions accurately reflect textual semantics, focusing on key action recognition and left-right directionality.
    
    \item \textbf{Temporal Coherence}
    evaluates whether multi-step motions occur in the correct causal order and are completed fully, without abrupt truncation or discontinuity.
    
\end{itemize}

\subsection{Motion-to-Text Evaluation}

While traditional metrics such as BLEU~\cite{Bleu} and CIDEr~\cite{Cider} are widely adopted for M2T, they mainly capture lexical overlap and provide limited diagnosis of motion-language alignment. In open-world scenarios, motions can be described in diverse yet semantically equivalent ways. To address this, M2T evaluation uses 1,000 motion samples with 272-dimensional representations from the MotionMillion~\cite{MotionMillion} validation split. For a unified evaluation interface, we convert them to HumanML3D's 263-dimensional representation when required by baseline models.
We render motion sequences into videos using Blender and apply YOLOv11-Pose~\cite{yolo11} to distinguish left–right body joints.
Left limbs are colored green and right limbs red, providing clear spatial orientation for downstream VLM evaluation.
A VLM evaluator assigns 0-5 scores for each criterion by jointly analyzing the generated text, rendered motion video, and ground-truth caption. Evaluation focuses on three key aspects:

\begin{itemize}

\item \textbf{Detail Consistency}  
examines how well the output text captures fine-grained visual and motion details. 

\item \textbf{Semantic Alignment}  
evaluates whether the generated textual description accurately captures the key semantics of the original motion.

\item \textbf{Temporal Coherence}  
assesses whether the order and timing of actions match those in the video.
\end{itemize}

\subsection{Bidirectional Consistency Evaluation}

While T2M and M2T individually evaluate generative and interpretive capabilities, they cannot determine whether a unified model preserves cross-modal semantics across bidirectional transformations. The correspondence between language and motion is inherently one-to-many. This intrinsic asymmetry makes it difficult to assess whether a model captures genuine cross-modal semantics rather than relying on superficial correlations. To address this challenge, Open-MoBench introduces a bidirectional consistency protocol that evaluates semantic preservation across cyclic motion-language transformations. We measure bidirectional consistency through two cyclic transformations:

\begin{equation}
\hat{T}' = f_{\text{M2T}}(f_{\text{T2M}}(T)), 
\end{equation}
\begin{equation}
\hat{M}' = f_{\text{T2M}}(f_{\text{M2T}}(M)),
\end{equation}
where $f_{\text{T2M}}(\cdot)$ and $f_{\text{M2T}}(\cdot)$ denote the T2M and M2T mappings, respectively. 
Here, $T$ and $M$ represent the input text and motion sequence, while $\hat{T}'$ and $\hat{M}'$ denote their reconstructed outputs obtained after a full transformation cycle. 

A unified model demonstrates strong bidirectional consistency when the regenerated text $\hat{T}'$ retains the semantics of the original description $T$, and the reconstructed motion $\hat{M}'$ preserves the motion semantics and temporal structure of the input $M$. We use a VLM to assess semantic consistency across input, intermediate, and reconstructed modalities, with scores ranging from 0–5 to indicate the level of semantic preservation. This protocol shifts evaluation from single-direction generation to cross-modal semantic preservation, allowing Open-MoBench to assess whether a unified model maintains motion-language consistency across cyclic transformations.

\section{Experiments and Results}

\begin{table*}[htbp]
\centering
\caption{Quantitative results of the T2M task on the MotionMillion dataset. The best scores are highlighted in bold, and the second-best scores are underlined. Results marked with $\triangledown$ denote variants trained with an additional symmetric M2T reward that mirrors the T2M reward design.}
\label{tab:t2m_results}
\setlength{\tabcolsep}{13pt}
\begin{tabular}{lcccccc}
\toprule
\textbf{Methods} & \multicolumn{3}{c}{\textbf{R-Precision}↑} & \textbf{FID}↓ & \textbf{MM-Dist}↓ & \textbf{Diversity}↑ \\
\cmidrule(lr){2-4}
& \textbf{Top1} & \textbf{Top2} & \textbf{Top3} &  &  \\
\midrule
ScaMo~\cite{Scamo} & 0.670 & 0.810 & 0.870 & 89.000 & - & -\\
GoToZero~\cite{MotionMillion}& 0.790 & 0.905 & 0.944 & 10.568 & 24.574 & 46.272\\
Open-UniMo SFT (w/o CoT) & 0.786 & 0.894 & 0.933 & 10.909 & 24.281 & 46.385\\
Open-UniMo SFT+GRPO (w/o CoT)$^{\triangledown}$ & 0.812 & 0.915 & 0.950 & \textbf{10.444}  & 23.628  & 46.457\\
Open-UniMo SFT+GRPO (w/o CoT) & 0.814 & 0.917 & 0.951 & \underline{10.526} & 23.555 & \underline{46.495}\\
Open-UniMo SFT& 0.782 & 0.892 & 0.932 & 11.159 & 24.332 & 46.304\\
Open-UniMo SFT+GRPO $^{\triangledown}$ & \underline{0.849} &  \underline{0.945} & \underline{0.971} & 19.909  & \underline{22.283} & 46.404\\
\rowcolor{blue!15}
\textbf{Open-UniMo SFT+GRPO} & \textbf{0.860} & \textbf{0.950} & \textbf{0.973} & 17.353 & \textbf{21.947} & \textbf{46.546} \\
\bottomrule
\end{tabular}
\end{table*}

\begin{table*}[htbp]
\centering
\caption{Quantitative results of the M2T task on the M2T subset of Open-MoBench. The best scores are highlighted in bold, and the second-best scores are underlined. Results marked with $\triangledown$ denote variants trained with an additional symmetric M2T reward that mirrors the T2M reward design. Results marked with $\lozenge$ indicate evaluation on the full validation set of the MotionMillion dataset.}
\label{tab:m2t_results}
\setlength{\tabcolsep}{14pt}
\begin{tabular}{lccccc}
\toprule
\textbf{Methods} & \textbf{BLEU-1} & \textbf{BLEU-4} & \textbf{ROUGE-L} & \textbf{CIDEr} & \textbf{BertScore} \\
\midrule
TM2T~\cite{TM2T}& 28.60 & 2.46 & 27.34 & 3.59 & 27.20 \\
MotionGPT~\cite{MotionGPT}& 34.69 & 4.33 & 25.01 & 5.53 & 31.81 \\
MotionAgent~\cite{MotionAgent}& 36.77 & 5.75 & 27.86 & 6.73 & 35.96 \\
Open-UniMo SFT (w/o CoT)& 52.80 & 11.78 & 31.96 & 10.51 & 40.29 \\
Open-UniMo SFT+GRPO (w/o CoT)$^{\triangledown}$ & 51.22 & 10.80 & 31.60 & 9.55 & 39.68 \\
Open-UniMo SFT+GRPO (w/o CoT) & 52.63 & 11.97 & 32.35 & 11.48 & 40.72 \\
Open-UniMo SFT& \underline{61.98} & \underline{20.48} & \underline{39.71} & \underline{25.62} & \underline{51.72}  \\
Open-UniMo SFT+GRPO$^{\triangledown}$ & 59.02 & 18.24 &  38.44 & 19.43  &  49.96 \\
\rowcolor{blue!15}
\textbf{Open-UniMo SFT+GRPO} & \textbf{64.44}  & \textbf{21.25} &  \textbf{40.17}  & \textbf{27.74}  &  \textbf{51.84}  \\
\midrule
Open-UniMo SFT$^{\lozenge}$ & 54.22 & 12.42  & 32.87 & 12.45 & 44.56 \\
\rowcolor{blue!15}
\textbf{Open-UniMo SFT+GRPO$^{\lozenge}$} & \textbf{57.56} &  \textbf{13.85} &   \textbf{33.69} &   \textbf{14.90} &  \textbf{45.97} \\
\bottomrule
\end{tabular}
\end{table*}

\subsection{Dataset}

We conduct experiments on the HumanML3D~\cite{T2M} and MotionMillion~\cite{MotionMillion} datasets. HumanML3D contains 14,616 motion clips collected from AMASS~\cite{AMASS} and HumanAct12~\cite{HumanAct12}, paired with 44,970 natural-language descriptions. 
MotionMillion is a million-scale open-source human motion-language corpus.
It comprises over 2,000 hours of human motion recordings and more than 2 million text-motion pairs, captured at a frame rate of 30~FPS. 
The dataset is curated from large-scale videos collected from diverse online sources, covering a wide range of real-world scenarios. 
In terms of motion captions, MotionMillion uses a two-step process. First, GPT-4o generates captions for video clips with prompts that include cues about age, emotion, movement style, and environment. Then, Llama~3.1-8B~\cite{llama3} paraphrases each caption $20$ times to enrich semantic diversity.
We further construct CoT annotations for the training pairs in both datasets.

\subsection{Evaluation Metrics}

\paragraph{T2M and M2T Metrics}
The T2M evaluation adopts metrics including FID, R-Precision (Top-1/2/3), and MM-Dist from~\cite{T2M} to measure the distributional similarity and retrieval capability.
In the M2T task, conventional linguistic metrics such as BLEU-1/4~\cite{Bleu}, ROUGE-L~\cite{Rouge}, CIDEr~\cite{Cider}, and BertScore~\cite{Bertscore} are employed to evaluate the quality of textual descriptions generated from motion inputs.

\paragraph{Open-MoBench Evaluation}
Beyond traditional metrics, Open-MoBench introduces a comprehensive evaluation framework that jointly assesses three key aspects using Gemini-3-Flash. For the T2M task, Open-MoBench provides VLM-guided assessments across three diagnostic dimensions: physical plausibility, semantic alignment, and temporal coherence.
For the M2T task, it measures detail consistency, semantic alignment, and temporal coherence of generated descriptions.
In addition, a bidirectional consistency protocol evaluates semantic preservation across cyclic T2M and M2T transformations, serving as a unified benchmark for open-world motion-language generation and understanding.

\subsection{Implementation Details}

The training of Open-UniMo comprises two stages: SFT and RL with GRPO. All experiments are conducted on 8 $\times$ A100 GPUs using the Qwen2.5-3B-Instruct~\cite{qwen2_5} model as the base language model.
In the SFT stage, we adopt the AdamW optimizer with a cosine annealing schedule and an initial learning rate of $2\times10^{-4}$. Open-UniMo is trained for a total of 250k iterations with a global batch size of 96, which takes about $18$ days. 

Then, GRPO is applied for 5{,}000 steps with a learning rate of $4\times10^{-5}$. Each iteration generates $G=6$ outputs with the global batch size of 60. The algorithm employs a clipping coefficient $\epsilon=0.2$ and a regularization factor $\beta=0.01$ to maintain stability and control exploration.

\begin{table*}[ht]
  \centering
  \caption{Quantitative results of the T2M task on the HumanML3D dataset. Each evaluation is repeated 20 times with average metrics and 95\% confidence intervals. The best scores are highlighted in bold, and the second-best scores are underlined.}
  \label{tab:unimo_t2m_compare}
  \setlength{\tabcolsep}{9pt}
  \begin{tabular}{lccccccc}
    \toprule
    \textbf{Methods} & \multicolumn{3}{c}{\textbf{R-Precision}↑} & \textbf{FID}↓ & \textbf{MM-Dist}↓ & \textbf{Diversity}↑ & \textbf{MModality}↑ \\
    \cmidrule(lr){2-4}
    & \textbf{Top1} & \textbf{Top2} & \textbf{Top3} &  &  \\
    \midrule
    MDM~\cite{MDM} & $0.320^{\pm 0.005}$ & $0.498^{\pm 0.004}$ & $0.611^{\pm 0.007}$ & $0.544^{\pm 0.044}$ & $5.566^{\pm 0.027}$ & $9.559^{\pm 0.086}$ & \textbf{2.799}$^{\pm 0.074}$  \\
    MLD~\cite{MLD} & $0.481^{\pm 0.003}$ & $0.673^{\pm 0.003}$ & $0.772^{\pm 0.002}$ & $0.473^{\pm 0.013}$ & $3.196^{\pm 0.010}$ & $9.724^{\pm 0.082}$ & $2.413^{\pm 0.072}$ \\
    MotionDiffuse~\cite{MotionDiffuse} & $0.491^{\pm 0.001}$ & $0.681^{\pm 0.001}$ & $0.782^{\pm 0.001}$ & $0.630^{\pm 0.001}$ & $3.113^{\pm 0.001}$ & $9.410^{\pm 0.049}$ & $1.553^{\pm 0.064}$ \\
    \midrule
    T2M~\cite{T2M}& $0.457^{\pm 0.002}$ & $0.559^{\pm 0.007}$ & $0.740^{\pm 0.003}$ & $1.067^{\pm 0.002}$ & $3.340^{\pm 0.008}$ & $9.188^{\pm 0.002}$ & $2.090^{\pm 0.088}$ \\
    TM2T~\cite{TM2T} & $0.424^{\pm 0.003}$ & $0.618^{\pm 0.003}$ & $0.729^{\pm 0.002}$ & $1.501^{\pm 0.017}$ & $3.467^{\pm 0.011}$ & $8.589^{\pm 0.076}$ & \underline{2.424}$^{\pm 0.079}$ \\
    T2M-GPT~\cite{T2M-GPT} & $0.491^{\pm 0.003}$ & $0.680^{\pm 0.003}$ & $0.775^{\pm 0.002}$ & \underline{0.116}$^{\pm 0.004}$ & $3.118^{\pm 0.011}$ & $9.761^{\pm 0.081}$ & $1.856^{\pm 0.111}$ \\
    MotionGPT~\cite{MotionGPT} & $0.492^{\pm 0.003}$ & $0.681^{\pm 0.003}$ & $0.778^{\pm 0.002}$ & $0.232^{\pm 0.008}$ & $3.096^{\pm 0.008}$ & $9.528^{\pm 0.071}$ & $2.008^{\pm 0.083}$ \\
    MoMask~\cite{Momask} & \underline{0.521}$^{\pm 0.002}$ & $0.713^{\pm 0.002}$ & $0.807^{\pm 0.002}$ & \textbf{0.045}$^{\pm 0.002}$ & $2.958^{\pm 0.008}$ & $9.620^{\pm 0.064}$ & $1.241^{\pm 0.064}$ \\
    MotionChain~\cite{MotionChain} & 0.504$^{\pm 0.003}$ & 0.617$^{\pm 0.002}$ & 0.790$^{\pm 0.003}$ & 0.248$^{\pm 0.009}$ & 3.033$^{\pm 0.010}$ & 9.470$^{\pm 0.075}$ & 1.727$^{\pm 0.014}$ \\
    MotionAgent~\cite{MotionAgent} & 0.515$^{\pm 0.004}$ & 0.691$^{\pm 0.003}$ & 0.801$^{\pm 0.004}$ & 0.230$^{\pm 0.009}$ & 2.967$^{\pm 0.020}$ & 9.908$^{\pm 0.102}$ & 2.142$^{\pm 0.014}$ \\
    MotionGPT-2~\cite{Motiongpt2}& 0.496$^{\pm 0.002}$ & 0.691$^{\pm 0.003}$ & 0.782$^{\pm 0.004}$ & 0.191$^{\pm 0.004}$ & 3.080$^{\pm 0.013}$ & 9.860$^{\pm 0.026}$ & 2.137$^{\pm 0.022}$ \\
    Motion-R1~\cite{MotionR1} & 0.515$^{\pm 0.003}$ & \underline{0.719}$^{\pm 0.002}$ & \underline{0.818}$^{\pm 0.002}$ & 0.201$^{\pm 0.004}$ & \underline{2.854}$^{\pm 0.010}$ & \underline{10.026}$^{\pm 0.075}$ & 2.317$^{\pm 0.105}$ \\
    \rowcolor{blue!15}
    \textbf{UniMo (Ours)}~\cite{unimo} & \textbf{0.539}$^{\pm 0.003}$ & \textbf{0.738}$^{\pm 0.002}$ & \textbf{0.831}$^{\pm 0.002}$ & 0.177$^{\pm 0.004}$ & \textbf{2.768}$^{\pm 0.010}$ & \textbf{10.042}$^{\pm 0.076}$ & 1.924$^{\pm 0.080}$\\
    \bottomrule
  \end{tabular}
\end{table*}

\begin{table}[ht]
\centering
\caption{Quantitative results of the M2T task on the HumanML3D dataset. The best scores are highlighted in bold, and the second-best scores are underlined.}
\label{tab:unimo_m2t_compare}
\setlength{\tabcolsep}{7pt}
\begin{tabular}{lccccc}
\toprule
\textbf{Methods} & \textbf{B@1↑} & \textbf{B@4↑} & \textbf{R-L↑} & \textbf{CIDEr↑} & \textbf{Bert↑} \\
\midrule
TM2T~\cite{TM2T} & 48.9 &  8.27 & 38.1 & 15.8 & 32.2 \\
LaMPM2T~\cite{Lamp} & 47.8 & 13.04 & 37.1 & 28.9 & 32.7 \\
MoTe~\cite{MoTe} & 46.7 & 11.15 & 37.4 & 31.5 & 30.3 \\
MotionGPT~\cite{MotionGPT} & 48.2 & 12.47 & 37.4 & 29.2 & 32.4 \\
MotionGPT-2~\cite{Motiongpt2} & 48.7 & 13.80 & 37.6 & 29.8 & 32.6 \\
MotionChain~\cite{MotionChain} & 48.1 & 12.56 & 33.9 & \underline{33.7} & 36.9 \\
MotionAgent~\cite{MotionAgent} & \underline{54.5} & \underline{17.65} & \underline{48.7} & \underline{33.7} & \underline{42.6} \\
\rowcolor{blue!15}
\textbf{UniMo (Ours)}~\cite{unimo}& \textbf{63.1} & \textbf{19.74} & \textbf{48.8} & \textbf{46.7} & \textbf{54.3} \\
\bottomrule
\end{tabular}
\end{table}

\begin{table*}[!htbp]
\centering
\caption{Ablation study on the effectiveness of CoT and different GRPO reward components on the full MotionMillion validation set.}
\label{tab:unimo_ablation_cot_reward}
\setlength{\tabcolsep}{2mm}
\begin{tabular}{ccccccccccccccc}
\toprule
\multirow{2}*{\textbf{CoT}} & \multirow{2}*{\textbf{$r_{\text{motion}}$}} & \multirow{2}*{\textbf{$r_{\text{semantic}}$}} & \multirow{2}*{\textbf{$r_{\text{format}}$}}
& \multicolumn{3}{c}{\textbf{R-Precision}↑} 
& \multirow{2}*{\textbf{FID}$\downarrow$} & \multirow{2}*{\textbf{MM-Dist}$\downarrow$} & \multirow{2}*{\textbf{Diversity}$\uparrow$}
& \multirow{2}*{\textbf{B@1}$\uparrow$} & \multirow{2}*{\textbf{B@4}$\uparrow$}& \multirow{2}*{\textbf{R-L}$\uparrow$} & \multirow{2}*{\textbf{CIDEr}$\uparrow$} & \multirow{2}*{\textbf{Bert}$\uparrow$} \\
\cmidrule(lr){5-7}
& & & &  \textbf{Top1} & \textbf{Top2} & \textbf{Top3} & & & & & & & & \\
\midrule
 &  &  &  & 0.786 & 0.894 & 0.933 & 10.909 & 24.281 & 46.385 & 53.85 & 12.27 & 32.67 & 11.73  & 44.04\\
 & $ \checkmark $ & $\checkmark$ &  & 0.809 & 0.914 & 0.949 & \textbf{10.440}  & 23.673& \textbf{46.789} & 54.83 & 12.82 & 33.19 & 12.93 & 44.67\\
  & $ \checkmark $ &  & $ \checkmark $ & 0.807 & 0.915 & 0.949 & 10.623  & 23.775& 46.195 & 54.83  & 12.64 & 33.01 & 12.69 & 44.53\\
 &  & $ \checkmark $ & $ \checkmark $ & 0.797 & 0.906 & 0.944 & 10.672 & 23.940 &  46.504   & 54.68 & 12.67 & 33.04 & 12.56 & 44.50\\
 & $ \checkmark $ & $ \checkmark $ & $ \checkmark $ &0.814 & 0.917 & 0.951 & \underline{10.526} & 23.555 & 46.495 & 55.02 & 12.85 & 33.17 & 12.87 &  44.61\\
 $ \checkmark $ &  &  &  &0.782 & 0.892 & 0.932 & 11.159 & 24.332 & 46.304& 54.22 &12.42  & 32.87 & 12.45 & 44.56\\
  $ \checkmark $ & $ \checkmark $ & $ \checkmark $ &  &  \underline{0.850} & \underline{0.946} & 0.970 & 19.208 & 22.286 &46.391   & 55.18 & 12.77 & 33.21 & 13.78 & 44.76\\
 $ \checkmark $ & $ \checkmark $ &  & $ \checkmark $ & 0.808 & 0.919 & 0.953 & 20.487 & 23.381 & 46.522  & \underline{57.19}  & \underline{13.64} &\textbf{33.83}  & \underline{14.54} & \underline{45.62}\\
 $ \checkmark $ &  & $ \checkmark $ & $ \checkmark $ & \underline{0.850} & \underline{0.946} & \underline{0.972} & 21.147 & \underline{22.266} & 46.271 & 56.21  & 13.30 & 33.25  & 13.70 &  45.05\\
 \rowcolor{blue!15}
 $ \checkmark $ & $ \checkmark $ & $ \checkmark $ & $ \checkmark $ &  \textbf{0.860} & \textbf{0.950} & \textbf{0.973} & 17.353 & \textbf{21.947} & \underline{46.546} &  \textbf{57.56} &  \textbf{13.85} &   \underline{33.69} &   \textbf{14.90} &  \textbf{45.97}\\
\bottomrule
\end{tabular}
\end{table*}

\begin{table*}[t]
\centering
\caption{Ablation study of the synergy effect of unified modeling on the HumanML3D and MotionMillion Datasets. All experiments on HumanML3D are conducted with CoT. Results on MotionMillion additionally compare settings with and without CoT.}
\label{tab:unimo_ablation_unified_all}

\begin{minipage}[t]{0.48\textwidth}
\centering
\small
{\footnotesize (a) HumanML3D: Effect of Unified Modeling on T2M Generation.}\vspace{5pt}

\resizebox{\linewidth}{!}{%
\begin{tabular}{llcccccc}
\toprule
\textbf{Train Task} & \textbf{Stage} & \multicolumn{3}{c}{\textbf{R-Precision}$\uparrow$} & \textbf{FID}$\downarrow$ & \textbf{MM-Dist}$\downarrow$ & \textbf{Div}$\uparrow$ \\
\cmidrule(lr){3-5}
 & & \textbf{Top1} & \textbf{Top2} & \textbf{Top3} & & & \\
\midrule
T2M      & SFT     & 0.438 & 0.613 & 0.702 & \underline{0.201} & 3.588 & 9.748 \\
T2M      & SFT+RL  & \underline{0.529} & \textbf{0.739} & \textbf{0.832} & 0.203 & \textbf{2.743} & \underline{9.780}  \\
T2M+M2T  & SFT     & 0.460 & 0.642 & 0.735 & 0.292 & 3.360 & 9.726 \\
 \rowcolor{blue!15}
T2M+M2T  & SFT+RL  & \textbf{0.539} & \underline{0.738} & \underline{0.831} & \textbf{0.177}  & \underline{2.768} & \textbf{10.042} \\
\bottomrule
\end{tabular}%
}
\end{minipage}
\hfill
\begin{minipage}[t]{0.48\textwidth}
\centering
\small
{\footnotesize (b) HumanML3D: Effect of Unified Modeling on M2T Captioning.}\vspace{5pt}

\resizebox{\linewidth}{!}{%
\begin{tabular}{llccccc}
\toprule
\textbf{Train Task} & \textbf{Stage} & \textbf{B@1}$\uparrow$ & \textbf{B@4}$\uparrow$ & \textbf{R-L}$\uparrow$ & \textbf{CIDEr}$\uparrow$ & \textbf{Bert}$\uparrow$ \\
\midrule
M2T      & SFT     & 54.62 & 14.74 & 43.3 & 31.05 & 47.88 \\
M2T      & SFT+RL  & \underline{61.91} & \underline{18.89} & \underline{47.5} & \underline{42.13} & \underline{52.88} \\
T2M+M2T  & SFT     & 55.34 & 15.31 & 43.2 & 32.08 & 48.16 \\
 \rowcolor{blue!15}
T2M+M2T  & SFT+RL  & \textbf{63.10} & \textbf{19.74} & \textbf{48.8} & \textbf{46.69} & \textbf{54.26} \\
\bottomrule
\end{tabular}%
}
\end{minipage}

\vspace{1em}

\begin{minipage}[t]{0.48\textwidth}
\centering
\small
{\footnotesize (c) MotionMillion: Effect of Unified Modeling on T2M Generation under Different CoT Settings.}\vspace{5pt}

\resizebox{\linewidth}{!}{%
\begin{tabular}{llcccccccc}
\toprule
\textbf{Train Task} & \textbf{Stage} & \textbf{CoT} & \multicolumn{3}{c}{\textbf{R-Precision}$\uparrow$} & \textbf{FID}$\downarrow$ & \textbf{MM-Dist}$\downarrow$ & \textbf{Div}$\uparrow$ \\
\cmidrule(lr){4-6}
 & & & \textbf{Top1} & \textbf{Top2} & \textbf{Top3} & & & \\
\midrule
T2M      & SFT     &  & 0.791 & 0.902 & 0.941 & 12.165 & 24.155 & 46.539 \\
T2M      & SFT+RL  &  & 0.804 & 0.913 & 0.949 & \underline{10.671} & 23.826 &  \textbf{46.548} \\
T2M      & SFT     & $\checkmark$ & 0.775 & 0.890 & 0.931 & 10.891 & 24.543 & 46.309 \\
T2M      & SFT+RL  & $\checkmark$ & \underline{0.852} & \underline{0.947} &\underline{0.972} & 18.107 & \underline{22.251} & 46.061 \\
T2M+M2T  & SFT     &  & 0.786 & 0.894 & 0.933 & 10.909 & 24.281 & 46.385 \\
T2M+M2T  & SFT+RL  &  & 0.814 & 0.917 & 0.951 & \textbf{10.526} & 23.555 & 46.495 \\
T2M+M2T  & SFT     & $\checkmark$ & 0.782 & 0.892 & 0.932 & 11.159 & 24.332 & 46.304 \\
 \rowcolor{blue!15}
T2M+M2T  & SFT+RL  & $\checkmark$ & \textbf{0.860} & \textbf{0.950} & \textbf{0.973} & 17.353 & \textbf{21.947} & \underline{46.546} \\
\bottomrule
\end{tabular}%
}
\end{minipage}
\hfill
\begin{minipage}[t]{0.48\textwidth}
\centering
\small
{\footnotesize (d) MotionMillion: Effect of Unified Modeling on M2T Captioning under Different CoT Settings.}\vspace{5pt}

\resizebox{\linewidth}{!}{%
\begin{tabular}{llccccccc}
\toprule
\textbf{Train Task} & \textbf{Stage} & \textbf{CoT} & \textbf{B@1}$\uparrow$ & \textbf{B@4}$\uparrow$ & \textbf{R-L}$\uparrow$ & \textbf{CIDEr}$\uparrow$ & \textbf{Bert}$\uparrow$ \\
\midrule
M2T      & SFT     &  & 34.32 & 2.61 & 22.87 & 1.35 & 32.17 \\
M2T      & SFT+RL  &  & 36.26 & 2.73 & 21.66 & 1.56 & 32.19 \\
M2T      & SFT     & $\checkmark$ & 27.34 & 1.85 & 20.90 & 1.28 & 31.91 \\
M2T      & SFT+RL  & $\checkmark$ & 37.05 & 2.60 & 22.48 & 1.87 & 32.41 \\
T2M+M2T  & SFT     &  & 53.85 & 12.27 & 32.67 & 11.73 & 44.04 \\
T2M+M2T  & SFT+RL  &  & \underline{55.02} & \underline{12.85} & \underline{33.17} & \underline{12.87} & \underline{44.61} \\
T2M+M2T  & SFT     & $\checkmark$ & 54.22 & 12.42 & 32.87 & 12.45 & 44.56 \\
 \rowcolor{blue!15}
T2M+M2T  & SFT+RL  & $\checkmark$ & \textbf{57.56} & \textbf{13.85} & \textbf{33.69} & \textbf{14.90} & \textbf{45.97} \\
\bottomrule
\end{tabular}%
}
\end{minipage}
\end{table*}

\begin{table}[t]
\centering
\caption{
Ablation study on the impact of motion token vocabulary size for the M2T task on the full validation set of the MotionMillion dataset.
}
\label{tab:token_ablation}

\begin{tabular}{lccccc}
\toprule
\textbf{Vocabulary Size} 
& \textbf{B@1}$\uparrow$ 
& \textbf{B@4}$\uparrow$ 
& \textbf{R-L}$\uparrow$ 
& \textbf{CIDEr}$\uparrow$ 
& \textbf{Bert}$\uparrow$ \\
\midrule
65536 & 34.32 & 2.61 & 22.87 & \textbf{1.35} & 32.17 \\
16384 & 36.28 & 2.84 & \textbf{24.04} & 1.31 & \textbf{33.39} \\
4096  & 36.16 & \textbf{3.22} & 23.21 & 1.01 & 31.48 \\
2048  & 33.42 & 2.59 & 21.68 & 0.93 & 27.85 \\
512   & \textbf{37.95} & 2.95 & 20.92 & 1.20 & 29.67 \\
\bottomrule
\end{tabular}
\end{table}

\begin{table*}[htbp]
\centering
\caption{Quantitative results of the T2M task on the Open-MoBench. The best scores are highlighted in bold. \textbf{Phy. Plaus.}: Physical Plausibility (0-5); \textbf{Sem. Align.}: Semantic Alignment (0-5); \textbf{Temp. Coh.}: Temporal Coherence (0-5); \textbf{Avg. }: Average Score (0-5); \textbf{Bi. Consis.}: Bidirectional Consistency (0-5).}
\label{tab:t2m_results_Open-MoBench}
\begin{tabular}{lccc|c|c}
\toprule
\textbf{Methods} & \textbf{Phy. Plaus.} & \textbf{Sem. Align.} & \textbf{Temp. Coh.} & \textbf{Avg.} & \textbf{Bi. Consis.}\\
\midrule
GoToZero\cite{MotionMillion} & 3.141 & 1.903 & 2.197 & 2.414 & -\\
Open-UniMo SFT (w/o CoT) & 3.218 & 1.871 & 2.229  & 2.439 & 1.587\\
Open-UniMo SFT+GRPO (w/o CoT) & 3.189 & 1.912 & 2.272 & 2.458 & 1.686\\
\rowcolor{blue!15}
\textbf{Open-UniMo SFT+GRPO} & \textbf{3.281} & \textbf{1.990} & \textbf{2.312} &  \textbf{2.528} & \textbf{1.742}\\
\bottomrule
\end{tabular}
\end{table*}

\begin{table*}[htbp]
\centering
\caption{Quantitative results of the M2T task on the Open-MoBench. The best scores are highlighted in bold. \textbf{Detail Consis.}: Detail Consistency (0-5); \textbf{Sem. Align.}: Semantic Alignment (0-5); \textbf{Temp. Coh.}: Temporal Coherence (0-5); \textbf{Avg.}: Average Score (0-5); \textbf{Bi. Consis.}: Bidirectional Consistency (0-5).}
\label{tab:m2t_results_Open-MoBench}
\begin{tabular}{lccc|c|c}
\toprule
\textbf{Methods} & \textbf{Detail Consis.} & \textbf{Sem. Align.} & \textbf{Temp. Coh.} & \textbf{Avg.} & \textbf{Bi. Consis.} \\
\midrule
TM2T~\cite{TM2T} & 0.995 & 1.145 & 1.301 & 1.147 & 1.714\\
MotionGPT~\cite{MotionGPT} & 1.082 & 1.182 & 1.383 & 1.216 & 1.862\\
MotionAgent~\cite{MotionAgent} & 1.133 & 1.302 & 1.491 & 1.309 & 1.813\\
\midrule
Open-UniMo SFT (w/o CoT) & 2.346 & 2.976 & 3.096 & 2.806 & \textbf{2.369}\\
Open-UniMo SFT+GRPO (w/o CoT) & 2.433 & 3.052 & 3.233 & 2.906 & 2.362\\
\rowcolor{blue!15}
\textbf{Open-UniMo SFT+GRPO} & \textbf{2.439} & \textbf{3.077} & \textbf{3.308} & \textbf{2.941} & 2.288\\
\bottomrule
\end{tabular}
\end{table*}

\subsection{Main Results}

\paragraph{Scaling from HumanML3D to MotionMillion}
We first revisit the results of our conference version UniMo~\cite{unimo} on HumanML3D~\cite{T2M} and then examine whether the same unified motion-language modeling paradigm can scale to million-scale open-world data. As shown in Tabs.~\ref{tab:unimo_t2m_compare} and~\ref{tab:unimo_m2t_compare}, UniMo achieves state-of-the-art or competitive performance on HumanML3D across major metrics, validating the effectiveness of unified generation and understanding with structured reasoning on a controlled benchmark. Open-UniMo further extends this paradigm to MotionMillion~\cite{MotionMillion}, which contains substantially richer motion dynamics, more diverse language descriptions, and a larger-scale motion-token space. The results in Tabs.~\ref{tab:t2m_results} and~\ref{tab:m2t_results} show that Open-UniMo maintains strong performance under this more challenging setting, suggesting that the proposed LMLM design remains effective when scaled from HumanML3D to million-scale open-world data.

\paragraph{Text-to-Motion Results}
We evaluate Open-UniMo on the MotionMillion validation set and compare it with ScaMo~\cite{Scamo} and GoToZero~\cite{MotionMillion} under a compatible MotionMillion evaluation setting. As shown in Tab.~\ref{tab:t2m_results}, Open-UniMo achieves the best R-Precision and MM-Dist among all methods. When trained without CoT, applying GRPO improves Open-UniMo over SFT in both retrieval alignment and motion-text distance, showing that task-specific rewards effectively refine autoregressive motion-token generation.

The CoT-guided variant further improves semantic alignment in T2M. The final Open-UniMo SFT+GRPO model reaches 0.860/0.950/0.973 R-Precision and the lowest MM-Dist of 21.947, clearly outperforming the corresponding model trained without CoT. We also observe that the CoT-based model has a higher FID than the model trained without CoT. This does not contradict the alignment gains. Under the one-to-many nature of T2M, CoT encourages the model to select semantically consistent generation paths rather than only matching the global training distribution. As a result, retrieval-based alignment metrics improve substantially, while the generated distribution measured by FID may be reshaped.

\paragraph{Motion-to-Text Results}
For M2T, we first evaluate Open-UniMo on the M2T subset of Open-MoBench and compare it with representative open-source baselines, including TM2T~\cite{TM2T}, MotionGPT~\cite{MotionGPT}, and MotionAgent~\cite{MotionAgent}. Since these baselines are built on the HumanML3D representation, we convert Open-MoBench's samples to the compatible format for fair comparison. As shown in the upper part of Tab.~\ref{tab:m2t_results}, Open-UniMo substantially outperforms all baselines.

 For motion understanding, CoT further improves motion understanding by helping the model abstract and verbalize motion semantics. On the Open-MoBench subset, CoT-guided SFT increases BLEU-1 from 52.80 to 61.98 and CIDEr from 10.51 to 25.62 compared with SFT without CoT. This indicates that the intermediate reasoning trace helps decompose a high-dimensional motion sequence into ordered semantic components before producing the final caption. After GRPO, the final model achieves the best performance, reaching 64.44 BLEU-1 and 27.74 CIDEr. The $\lozenge$ rows in Tab.~\ref{tab:m2t_results} further show consistent gains from SFT to SFT+GRPO on the full MotionMillion validation set, confirming that the improvements are not limited to the Open-MoBench subset.

We additionally compare with a symmetric reward variant marked with $\triangledown$. In this variant, M2T is trained with an additional reward that mirrors the T2M alignment design: embeddings from the frozen encoders of~\cite{MotionMillion} are used to compute the similarity between the generated caption and both the ground-truth caption and the input motion. Its weaker performance shows that M2T should not simply inherit T2M-style motion-space alignment rewards. Since M2T requires compressing high-dimensional motion sequences into compact language semantics, caption-level semantic rewards are more suitable for optimizing motion understanding.

\subsection{Ablation Study}
We conduct ablation studies on MotionMillion to examine the effects of CoT and GRPO reward components. As shown in Tab.~\ref{tab:unimo_ablation_cot_reward}, adding GRPO rewards consistently improves the SFT baseline without CoT. With all reward components, the model improves R-Precision from 0.786/0.894/0.933 to 0.814/0.917/0.951, reduces MM-Dist from 24.281 to 23.555, and also improves M2T captioning metrics. These results show that GRPO provides effective post-training signals for autoregressive motion-language modeling through format control and cross-modal alignment rewards.
CoT provides a complementary semantic structure. CoT alone does not consistently improve all metrics, but combining CoT with GRPO leads to clear gains in both generation and understanding. The full model achieves the best R-Precision, MM-Dist, BLEU-1, BLEU-4, CIDEr, and BertScore, indicating that CoT is most effective when its intermediate reasoning is optimized together with task-specific rewards.

The ablation also reveals a difference between distributional fidelity and semantic alignment. Models trained without CoT can obtain lower FID, while CoT-based variants achieve stronger retrieval alignment and motion-text consistency. This suggests that CoT encourages semantics-guided generation, which may reshape the generated motion distribution while improving alignment to the input text. Overall, CoT and GRPO play complementary roles: CoT builds a shared semantic bridge between language and motion, while GRPO refines the model toward better structural correctness and cross-modal alignment.

\subsection{The Synergy Effect of Unified Modeling}
We further examine whether joint T2M and M2T training brings consistent benefits across dataset scales. As shown in Tab.~\ref{tab:unimo_ablation_unified_all}, unified training improves performance on HumanML3D for both directions, and the same trend remains on the larger MotionMillion dataset. This suggests that the synergy between generation and understanding is not limited to controlled benchmarks, but can also generalize to million-scale open-world data.
The effect is especially clear on MotionMillion. T2M-only training can already learn effective text-conditioned motion generation, and unified training further preserves or improves this ability, achieving the best R-Precision in the final setting. In contrast, M2T-only training remains weak across SFT, GRPO, CoT, and their combinations. This indicates that the M2T bottleneck is not simply due to the absence of reasoning traces or reward-based post-training, but is more closely related to the difficulty of extracting stable semantics from one-way motion-token sequences.

To further examine whether this bottleneck comes from motion-token capacity, we conduct a separate vocabulary-size ablation under the M2T-only SFT setting. As shown in Tab.~\ref{tab:token_ablation}, varying the vocabulary size from 512 to 65,536 does not produce a consistent performance trend: different metrics peak at different vocabulary sizes, and all results remain much lower than unified training. This shows that M2T-only performance is not determined by motion-token vocabulary size alone.
Unified training substantially alleviates this limitation. On MotionMillion, joint T2M+M2T SFT raises M2T BLEU-1 from 34.32 to 53.85 without CoT, while the full unified model with CoT and GRPO further reaches 57.56 BLEU-1 and 14.90 CIDEr. These results suggest that, in AR-based motion-language modeling, the learnable T2M path provides semantic grounding for motion tokens and helps the model acquire stronger cross-modal representations for M2T, demonstrating that generation facilitates understanding.

\subsection{Evaluation on Open-MoBench}

\paragraph{Text-to-Motion Evaluation}
Tab.~\ref{tab:t2m_results_Open-MoBench} reports the T2M evaluation results on Open-MoBench. Compared with GoToZero~\cite{MotionMillion}, Open-UniMo improves the average score from 2.414 to 2.528 and achieves higher scores in physical plausibility, semantic alignment, and temporal coherence. Among Open-UniMo variants, GRPO improves the model trained without CoT, while the final CoT-based SFT+GRPO model performs best across the main T2M dimensions.

\paragraph{Motion-to-Text Evaluation}
Tab.~\ref{tab:m2t_results_Open-MoBench} shows that Open-UniMo also achieves clear advantages on M2T evaluation. The final model obtains the highest average score of 2.941, compared with 1.309 from MotionAgent~\cite{MotionAgent}, with consistent improvements in detail consistency, semantic alignment, and temporal coherence. The gains over the model trained without CoT further indicate that CoT and GRPO help convert high-dimensional motion inputs into more coherent and semantically organized captions.

\paragraph{Bidirectional Consistency Evaluation}
Open-MoBench further reveals that bidirectional consistency is direction-dependent. In the T2M$\rightarrow$M2T cycle, the consistency score increases from 1.587 for Open-UniMo SFT (w/o CoT) to 1.742 for the final model. Since this cycle starts from text, CoT can expand the input into a structured motion plan before generation, making the synthesized motion easier to interpret and recover as text.

The M2T$\rightarrow$T2M cycle behaves differently. Although the final model achieves the best M2T score, it does not obtain the highest consistency score in this direction. We observe that CoT-guided M2T outputs are often more semantically organized, but the resulting captions may not always specify all details needed for reconstructing the original motion. 
Since the subsequent T2M step relies on this intermediate caption, this under-specification can lead to a reconstructed motion that is semantically plausible but different from the original motion, thereby reducing M2T$\rightarrow$T2M consistency. This result suggests that CoT improves semantic organization in M2T, but text-based intermediate representations may not fully preserve all motion details needed for faithful reconstruction.

\begin{figure*}[!htbp]
\centering
\includegraphics[width=\linewidth]{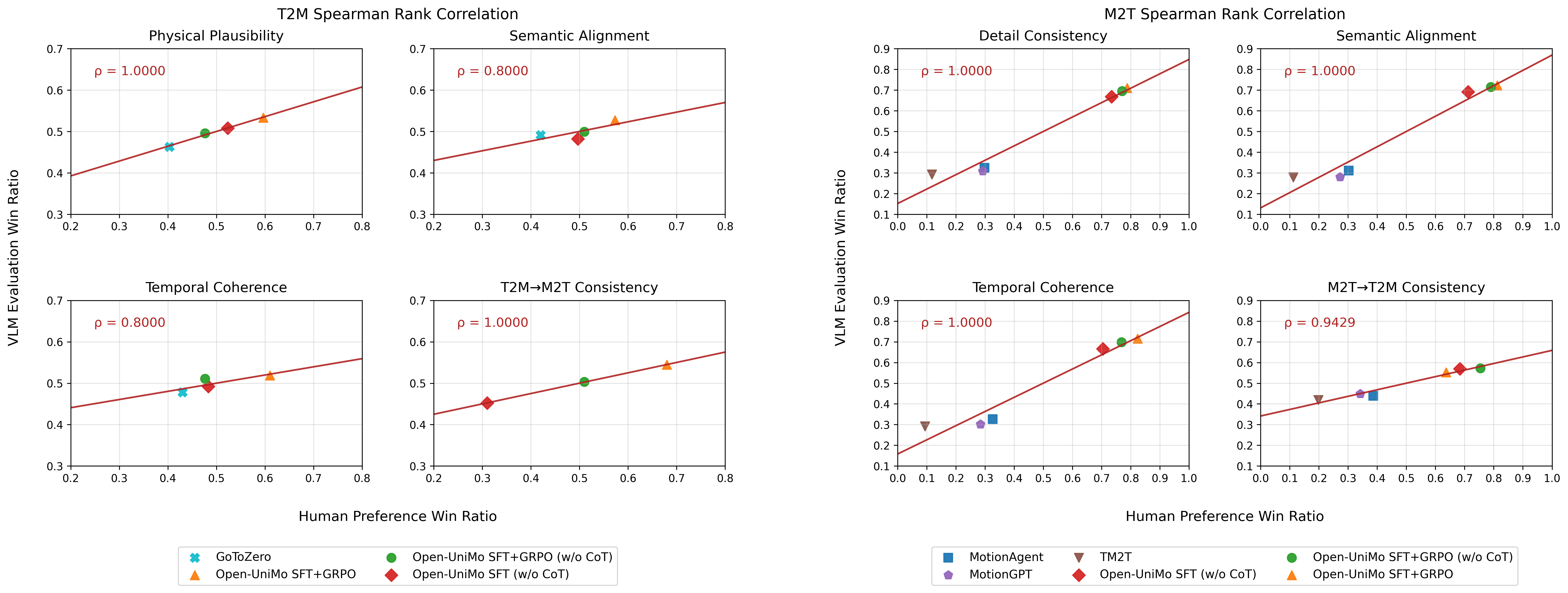}
\caption{
Spearman rank correlation between human preference and VLM-as-Judge evaluation on Open-MoBench. Each point represents one model, where the x-axis denotes the human preference win ratio and the y-axis denotes the VLM evaluation win ratio under the same task and evaluation dimension. The fitted line visualizes the ranking trend, and $\rho$ denotes the Spearman rank correlation between model-level human and VLM rankings.
}
\label{fig:human_vlm_alignment}
\end{figure*}

\subsection{Human Preference Testing}
\label{sec:human_preference_testing}
The goal of human preference testing is to verify whether the VLM-as-Judge scores used in Open-MoBench can serve as a reliable proxy for human perception. Following the validation protocol of VBench~\cite{vbench}, we compare model rankings induced by VLM scores with those obtained from human preferences. For each task and evaluation dimension, we randomly sample 100 examples from Open-MoBench and collect outputs from all compared models under the same input. We invite five volunteers with master's degrees to participate in the evaluation. All model identities are anonymized, and each comparison is conducted under a single specified criterion, such as semantic alignment, temporal coherence, or bidirectional consistency.

We convert human annotations into pairwise win ratios. For each pair of model outputs, the preferred one receives 1 and the other receives 0. Preferences are averaged over the five annotators; if the averaged preference difference is smaller than 0.05, the pair is treated as a tie and both models receive 0.5. VLM scores are converted into pairwise win ratios using the same model pairs and criteria, where the output with the higher VLM score is treated as preferred. We then compute the Spearman rank correlation between model-level human win ratios and VLM win ratios.

As shown in Fig.~\ref{fig:human_vlm_alignment}, VLM-as-Judge shows strong agreement with human preference. The rank correlations are perfect on all M2T dimensions and T2M$\rightarrow$M2T consistency, and remain high on M2T$\rightarrow$T2M consistency as well as the main T2M dimensions. These results indicate that Open-MoBench provides a reliable and scalable approximation to human evaluation for unified motion-language assessment.
Under the pairwise preference protocol, both human and VLM comparisons rank the full CoT+GRPO model below the non-CoT variants on M2T$\rightarrow$T2M consistency, which is consistent with our analysis that better semantic organization in M2T does not necessarily translate into more faithful motion reconstruction from the intermediate caption.

\section{Conclusion}
In this work, we introduced Open-UniMo, a unified Large Motion-Language Model for open-world T2M generation and M2T understanding. By extending Qwen's vocabulary with 64K motion tokens, Open-UniMo builds a unified token space that treats motion as a first-class modality and enables large-scale motion-language modeling on million-scale open-world data. We further incorporated CoT as a shared intermediate representation between language and motion, and developed a CoT-guided SFT$\rightarrow$GRPO training pipeline to improve semantic alignment and mitigate cumulative errors in autoregressive motion-token prediction.
To evaluate unified motion-language intelligence beyond conventional single-direction metrics, we proposed Open-MoBench, a VLM-guided benchmark that jointly assesses T2M generation, M2T understanding, and bidirectional consistency across diverse evaluation dimensions. Extensive experiments demonstrate that Open-UniMo achieves state-of-the-art performance on both traditional metrics and Open-MoBench. More importantly, our analysis reveals a key property of AR-based motion-language modeling: M2T understanding is not primarily constrained by motion-token vocabulary size, as T2M-only training can be effectively optimized whereas M2T-only training does not exhibit a consistent positive trend with vocabulary size. By coupling M2T with the learnable T2M generation path, unified training yields stronger cross-modal representations, demonstrating that generation can facilitate understanding in AR-based motion-language models. These results provide a strong step toward unified, scalable, and interpretable motion-language intelligence in open-world embodied AI.

\bibliographystyle{IEEEtran}
\bibliography{ref}

\twocolumn[
\begin{center}

{\Large\bfseries Supplementary Material\par}

\vspace{1.2em}

\end{center}
]
\setcounter{section}{0}
\section{Additional Details of Open-MoBench}

\begin{figure}[!htbp]
    \centering
    \includegraphics[width=0.7\linewidth]{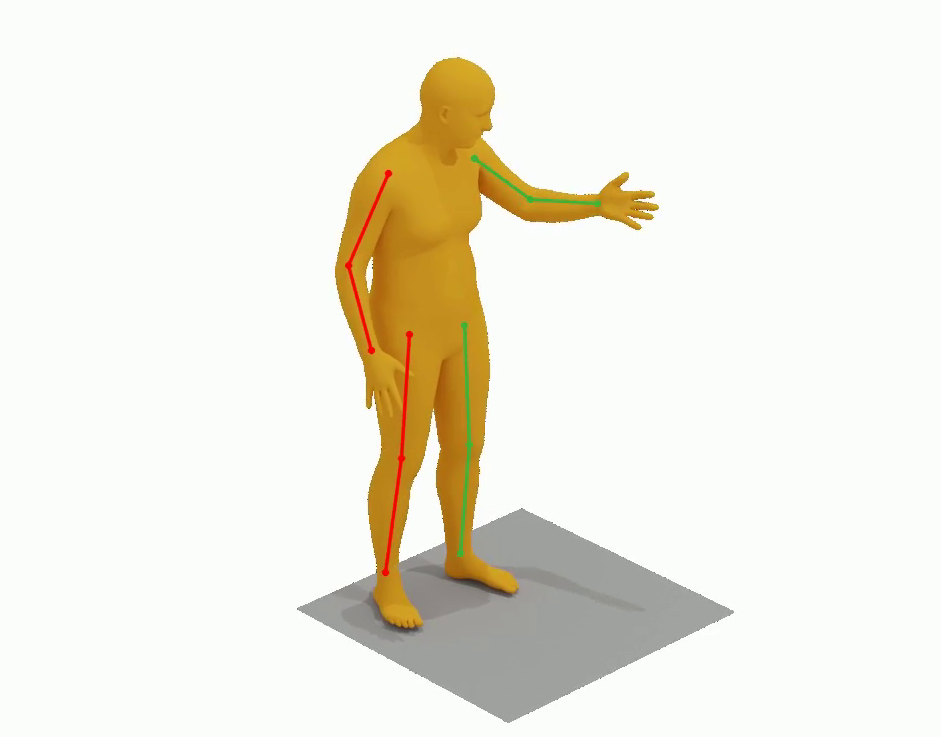}
    \caption{Left--right limb annotation for VLM-based motion evaluation. Blender-rendered motion frames are processed by YOLOv11-Pose~\cite{yolo11}, with {\color{ForestGreen}green} indicating left body limbs and {\color{red}red} indicating right body limbs.}
    \label{fig:yolo_pose}
\end{figure}

Open-MoBench evaluates motion-language models through rendered motion videos, which allows VLMs to assess human actions from visual evidence rather than raw kinematic sequences. Each motion sequence is first rendered in Blender and then processed by YOLOv11-Pose~\cite{yolo11}. The detected limbs are color-coded in the rendered frames, with green denoting the left body limbs and red denoting the right body limbs, as shown in Fig.~\ref{fig:yolo_pose}. This visual encoding makes directional and body-part-specific cues more explicit for VLM-based scoring.

\begin{figure}[!ht]
  \centering

  \includegraphics[width=\linewidth]{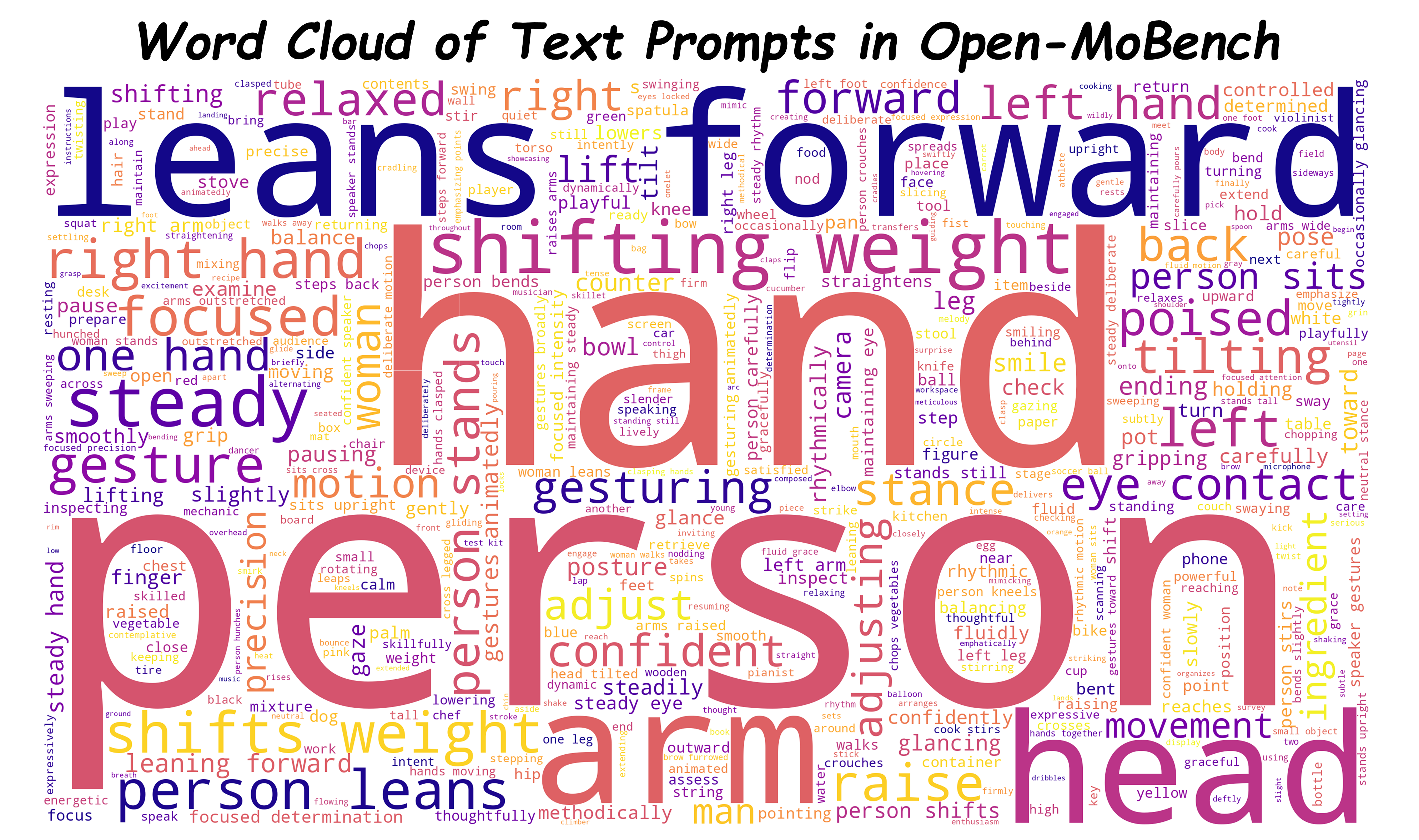}

  \vspace{0.8em}

  \includegraphics[width=\linewidth]{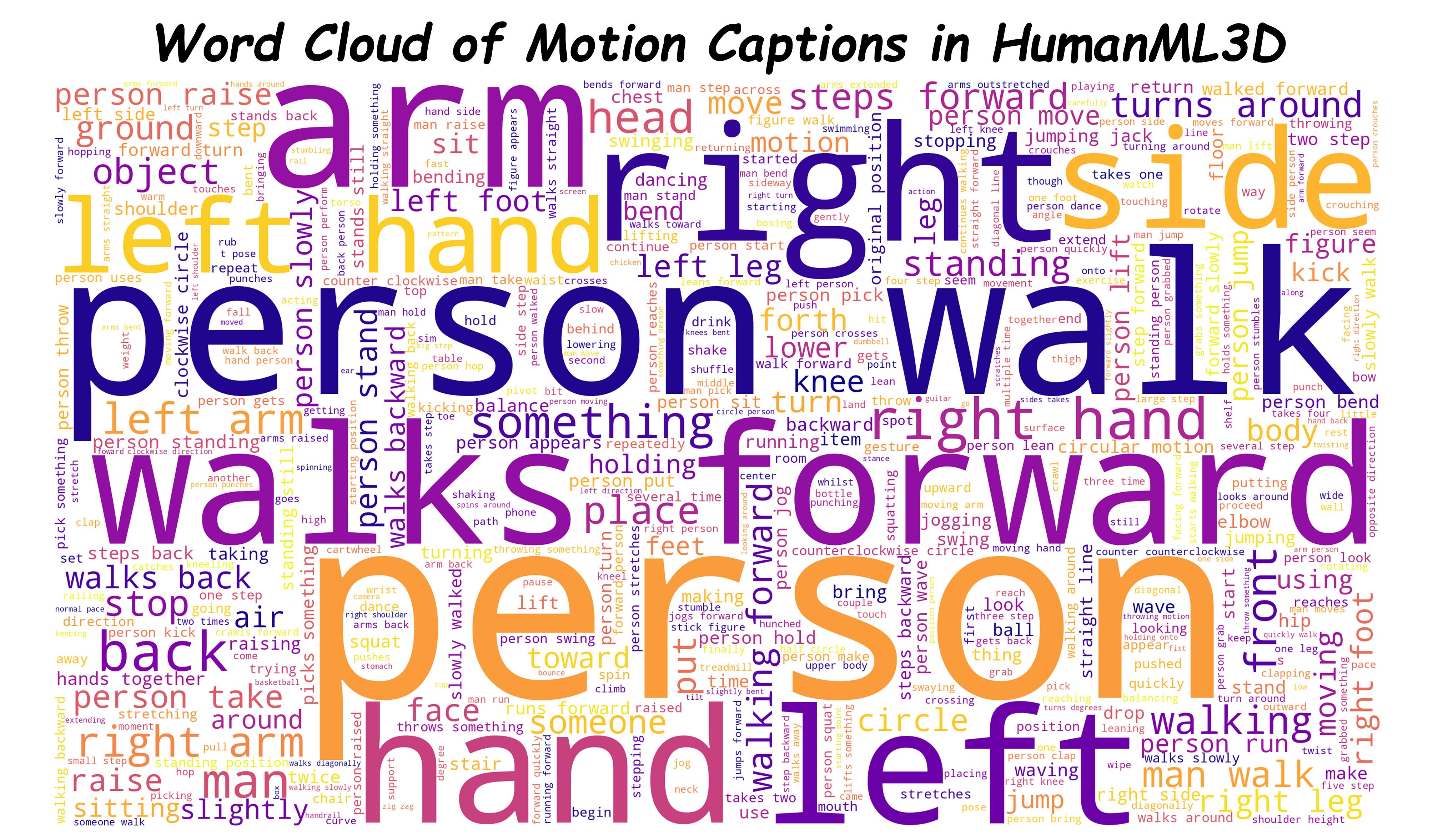}

  \caption{A comparative word cloud analysis of the T2M evaluation in Open-MoBench (top) and HumanML3D~\cite{T2M} (bottom).}
  \label{fig:wordcloud}
\end{figure}

Beyond the rendering protocol, Open-MoBench is designed to evaluate open-world linguistic coverage. Fig.~\ref{fig:wordcloud} compares the T2M prompts in Open-MoBench with motion captions from HumanML3D~\cite{T2M}. Compared with HumanML3D, Open-MoBench contains broader open-world action descriptions and richer linguistic diversity.

\begin{figure}[!ht]
\centering
\includegraphics[width=0.9\linewidth]{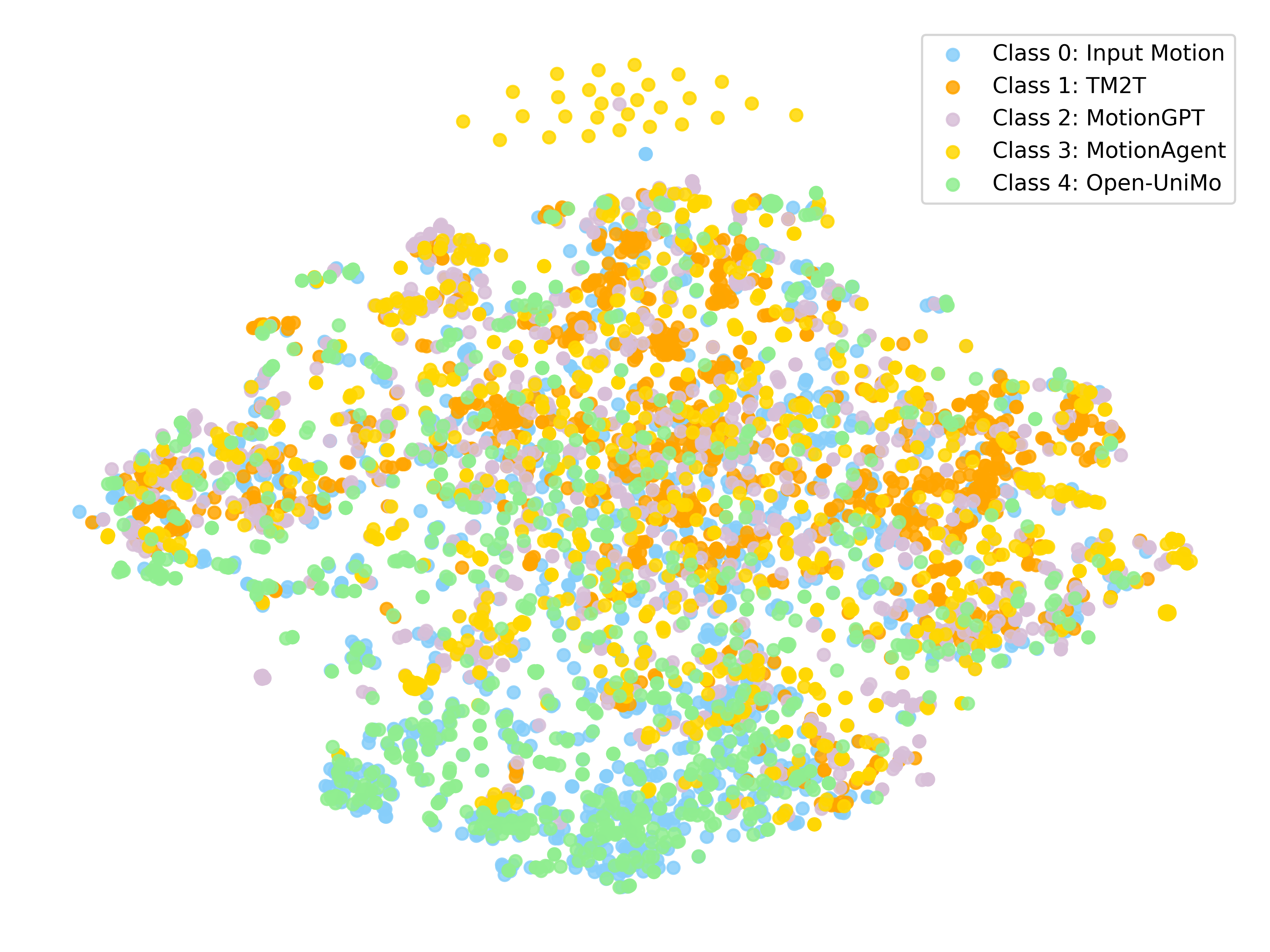}
\caption{t-SNE visualization of the M2T$\rightarrow$T2M bidirectional consistency evaluation in the motion feature space. Open-UniMo ({\color{OliveGreen}green}) shows stronger overlap with ground-truth clusters ({\color{RoyalBlue}blue}), suggesting better preservation of motion semantics.}
\label{fig:tsne}
\end{figure}

\begin{figure*}[!ht]
    \centering
    \includegraphics[width=\textwidth]{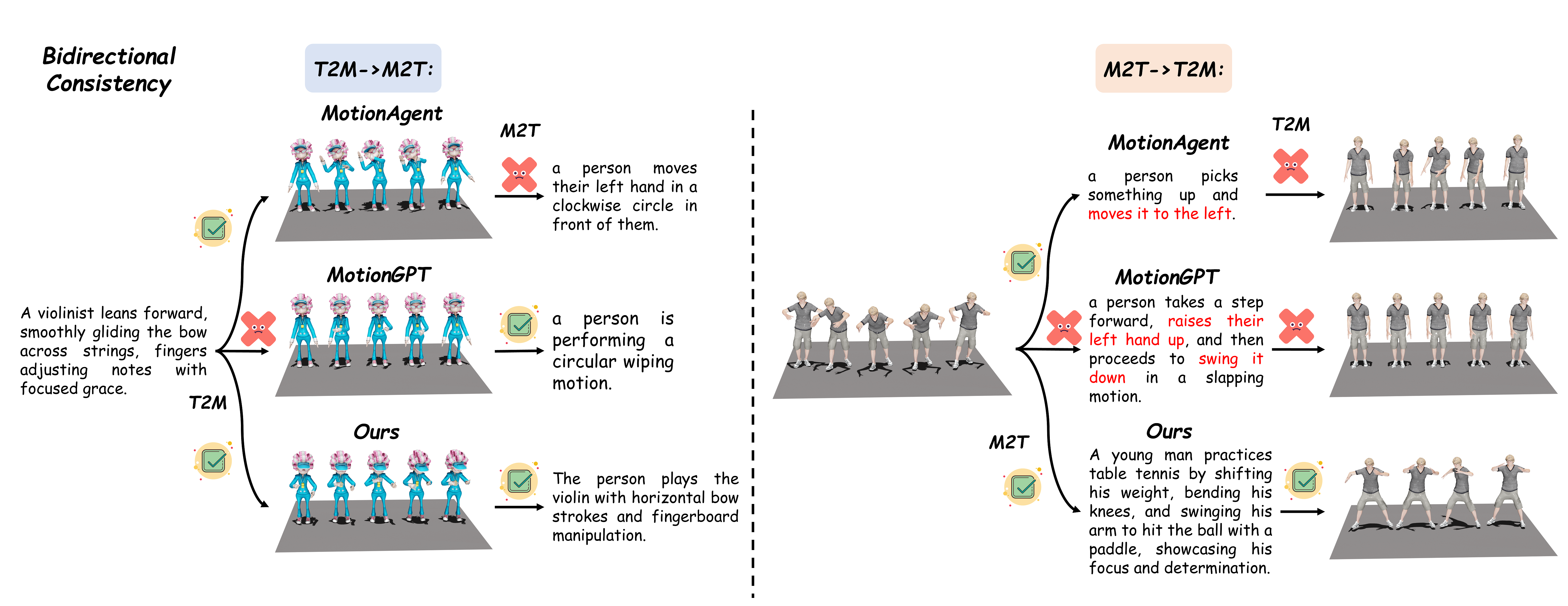}
    \caption{Qualitative comparison of our method with prior models \cite{MotionGPT, MotionAgent} on bidirectional consistency. Our approach better maintains semantic alignment between text and motion across both T2M and M2T directions.}
    \label{fig:Bidirectional}
\end{figure*}

Bidirectional consistency is analyzed through qualitative examples and motion-feature visualization. Fig.~\ref{fig:Bidirectional} compares Open-UniMo with prior methods~\cite{MotionGPT,MotionAgent} under T2M$\rightarrow$M2T and M2T$\rightarrow$T2M cycles, focusing on whether input semantics are preserved after a complete cross-modal transformation. For the M2T$\rightarrow$T2M cycle, Fig.~\ref{fig:tsne} visualizes motion features encoded by the motion encoder from~\cite{MotionMillion}. The stronger overlap between Open-UniMo and the ground-truth clusters suggests better preservation of the original motion semantics after the motion--text--motion cycle.

\begin{figure*}[!htbp]
    \centering
    \includegraphics[width=\textwidth]{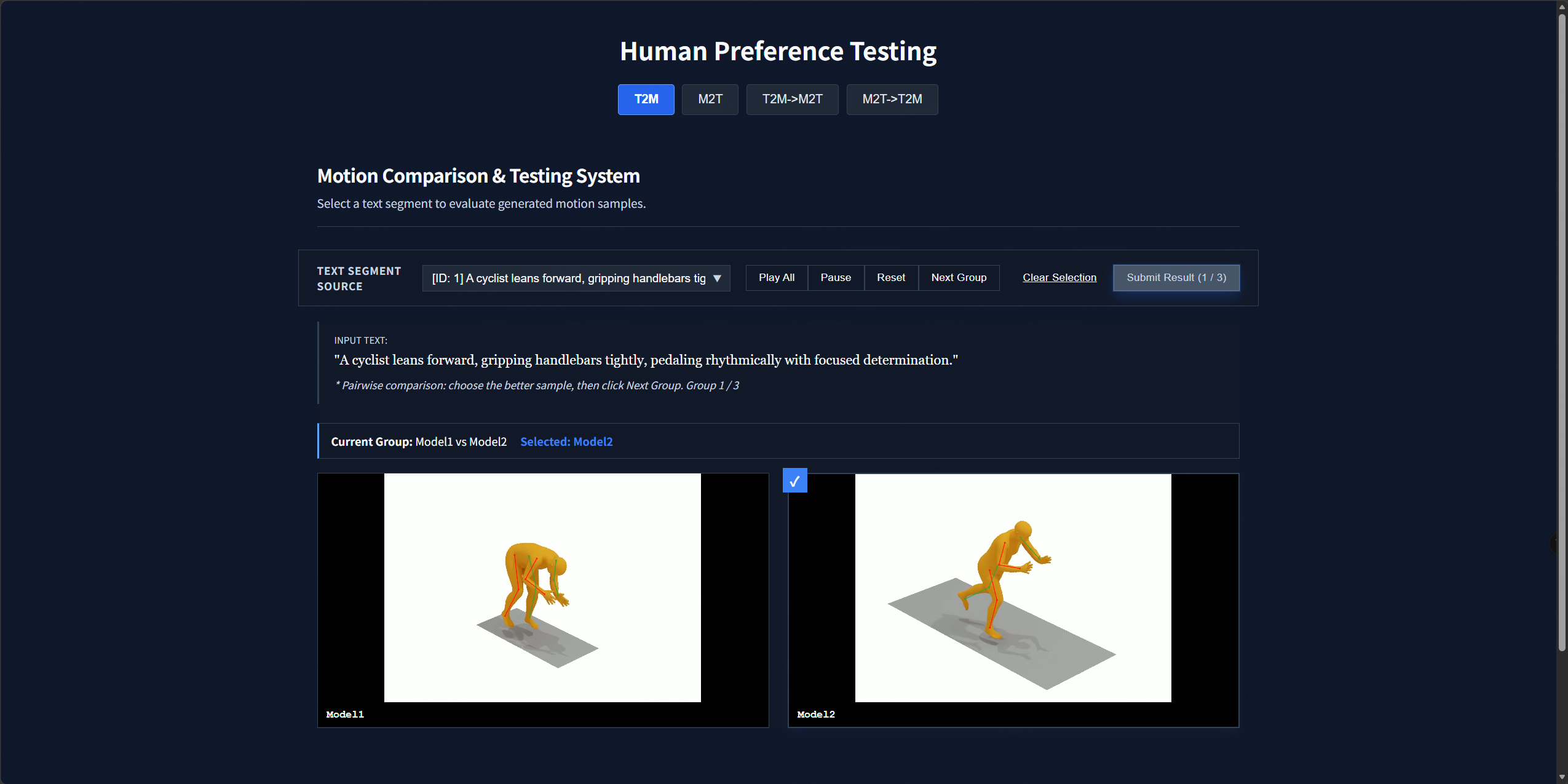}
    \caption{Human preference testing interface. Annotators compare anonymized model outputs under the same criterion and select the preferred result.}
    \label{fig:Human_Preferences_Testing_System}
\end{figure*}

The human preference study adopts a pairwise comparison interface, as shown in Fig.~\ref{fig:Human_Preferences_Testing_System}. For each sample group, annotators compare anonymized model outputs under the same criterion and select the preferred result.

\section{Prompt Templates}

Prompt design plays an important role in both CoT supervision and benchmark evaluation. Fig.~\ref{fig:cot_and_training_prompt} summarizes the prompts used for CoT annotation and T2M/M2T training. 
Fig.~\ref{fig:open_mobench_prompt} provides the VLM-as-Judge prompts used in Open-MoBench. These prompts specify the scoring criteria and output format for T2M generation, M2T understanding, and bidirectional consistency evaluation. For M2T$\rightarrow$T2M consistency, directly comparing two motion videos can be unreliable. Therefore, we compare the reconstructed motion video with the ground-truth caption of the input motion, using the caption as a textual proxy for the original motion semantics.

\begin{figure*}[!htbp]
    \centering
    \includegraphics[width=\textwidth]{img_sup/cot_and_training_prompt.png}
    \caption{Prompt templates for CoT annotation and model training. The prompts define structured reasoning and task-specific output formats for unified T2M generation and M2T understanding.}
    \label{fig:cot_and_training_prompt}
\end{figure*}

\begin{figure*}[!htbp]
    \centering
    \includegraphics[width=\textwidth]{img_sup/open_mobench_prompt.png}
    \caption{Prompt templates for Open-MoBench VLM-as-Judge evaluation, covering T2M generation, M2T understanding, and bidirectional consistency scoring.}
    \label{fig:open_mobench_prompt}
\end{figure*}


\vfill

\end{document}